%% file: main.tex
\documentclass[10pt,twocolumn,letterpaper]{article}

\usepackage[pagenumbers]{cvpr} 

\input{preamble}
\definecolor{cvprblue}{rgb}{0.21,0.49,0.74}
\usepackage[pagebackref,breaklinks,colorlinks,allcolors=cvprblue]{hyperref}
\usepackage{booktabs} 
\usepackage{graphicx} 
\usepackage{array}    
\usepackage{siunitx}  
\usepackage{etoolbox} 
\usepackage{xcolor}   
\usepackage{makecell}
\usepackage{float}
\usepackage{multirow}

\newcommand{\best}[1]{\textcolor{red}{\textbf{#1}}}

\usepackage{enumitem}     
\usepackage[T1]{fontenc}    
\definecolor{promptbg}{gray}{0.95}

\setlist{noitemsep, topsep=3pt, partopsep=0pt, parsep=2pt}

\input{preamble}
\definecolor{cvprblue}{rgb}{0.21,0.49,0.74}
\usepackage[pagebackref,breaklinks,colorlinks,allcolors=cvprblue]{hyperref}

\usepackage[left=2cm, right=2cm, top=2.5cm, bottom=2.5cm]{geometry}
\usepackage[dvipsnames]{xcolor} 
\usepackage[breakable]{tcolorbox} 
\usepackage{parskip} 

\newtcolorbox{promptbox}{
  colback=black!5,   
  colframe=black!20, 
  boxrule=0.5pt,     
  arc=2mm,           
  fontupper=\ttfamily, 
  left=3mm,          
  right=3mm,         
  top=2mm,           
  bottom=2mm,        
  breakable=true     
}

\def\paperID{701} 
\def\confName{CVPR}
\def\confYear{2026}

\title{I2I-Bench: A Comprehensive Benchmark Suite for Image-to-Image \\ Editing Models}
\author{Juntong Wang, Jiarui Wang, Huiyu Duan, Jiaxiang Kang, Guangtao Zhai, Xiongkuo Min\\
Institute of Image Communication and Network Engineering\\
Shanghai Jiao Tong University, Shanghai, China\\
{\tt\small \{wang13029187978\}@sjtu.edu.cn} 
}

\begin{document}
\maketitle

\input{sec/0_abstract}    
\input{sec/1_intro}
\input{sec/2_related_work}
\input{sec/3_I2IBench_Suite}
\input{sec/4_Experiment}
\input{sec/6_conclusion}
{
    \small
    \bibliographystyle{ieeenat_fullname}
    \bibliography{main}
}


\input{sec/Appendix}

\end{document}

%% file: sec/0_abstract.tex
\begin{abstract}
Image editing models are advancing rapidly, yet comprehensive evaluation remains a significant challenge. Existing image editing benchmarks generally suffer from limited task scopes, insufficient evaluation dimensions, and heavy reliance on manual annotations, which significantly constrain their scalability and practical applicability. To address this, we propose \textbf{I2I-Bench}, a comprehensive benchmark for image-to-image editing models, which features \textbf{(i) diverse tasks}, encompassing 10 task categories across both single-image and multi-image editing tasks, \textbf{(ii) comprehensive evaluation dimensions}, including 30 decoupled and fine-grained evaluation dimensions with automated hybrid evaluation methods that integrate specialized tools and large multimodal models (LMMs), and \textbf{(iii) rigorous alignment validation}, justifying the consistency between our benchmark evaluations and human preferences.
Using I2I-Bench, we benchmark numerous mainstream image editing models, investigating the gaps and trade-offs between editing models across various dimensions. We will open-source all components of I2I-Bench to facilitate future research.
\end{abstract}

%% file: sec/1_intro.tex
\section{Introduction}
\label{sec:intro}

Image editing has long been important tasks in computer vision \cite{wu2025qwenimagetechnicalreport,labs2025flux1kontextflowmatching,wu2025omnigen2,nanobanana}. Driven by the advancements of large multimodal models (LMMs) \cite{Qwen3VL,internvl3.5,gemma3,coreteam2025mimovltechnicalreport,gpt-4o,gemini2.5pro,li2023llava}, image editing methods have made rapid progress in the past few years, shifting from traditional local operations to instruction-induced content modification \cite{brooks2023instructpix2pix,icedit}.
 Moreover, the capabilities of editing models have also rapidly expanded, achieving not only single-image editing (SE) \cite{deng2025bagel,brooks2023instructpix2pix,icedit,gotedit,emuedit} but also more challenging multiple-image-editing (ME) \cite{nanobanana,xia2025dreamomni2multimodalinstructionbasedediting}. The rapid evolution across tasks and modalities underscores the urgent need for a new comprehensive benchmarks designed to rigorously assess these advanced capabilities.

However, existing evaluation paradigms exhibit clear shortcomings, struggling to keep pace with the rapid advancement of editing model capabilities. First, traditional metrics \cite{ssim,FID,clipscore} like PSNR or LPIPS \cite{lpips} are fundamentally inadequate, as they primarily assess pixel-level similarity and completely fail to evaluate whether complex semantic edits have been correctly executed. Though some more recent evaluation benchmarks \cite{xu2025lmm4edit,wang2025tdve,editbench,tedbench,basu2023editval,geval} have attempted to address this, they remain limited in both scope and granularity. Some benchmarks \cite{tedbench} possess a very limited test scope \cite{dover,fastvqa,lmm4lmm,clipscore,pickscore}, while others \cite{mps,blip2,fga,hpsv2,imagereward} lack the independent, quantitative analysis in terms of fine-grained dimensions which are crucial for editing model understanding and optimization. Furthermore, in terms of evaluation methods, current benchmarks face significant limitations. Some rely on manual annotation methods that are costly, non-scalable, and difficult-to-reproduce. Other approaches, such as score-based training methods \cite{xu2025lmm4edit}, still heavily rely on these manual annotations. Meanwhile, many automated methods perform poorly when handling complex semantic and cognitive tasks. More importantly, despite the emergence of multi-image editing task \cite{wu2025qwenimagetechnicalreport,xia2025dreamomni2multimodalinstructionbasedediting,nanobanana,wu2025omnigen2}, existing evaluation frameworks almost entirely lack coverage of such challenging tasks. A comprehensive and effective image editing evaluation suite is still lacking.

\begin{figure*}[t]
    \centering
    \vspace{-5pt}
    \includegraphics[width=1\linewidth]{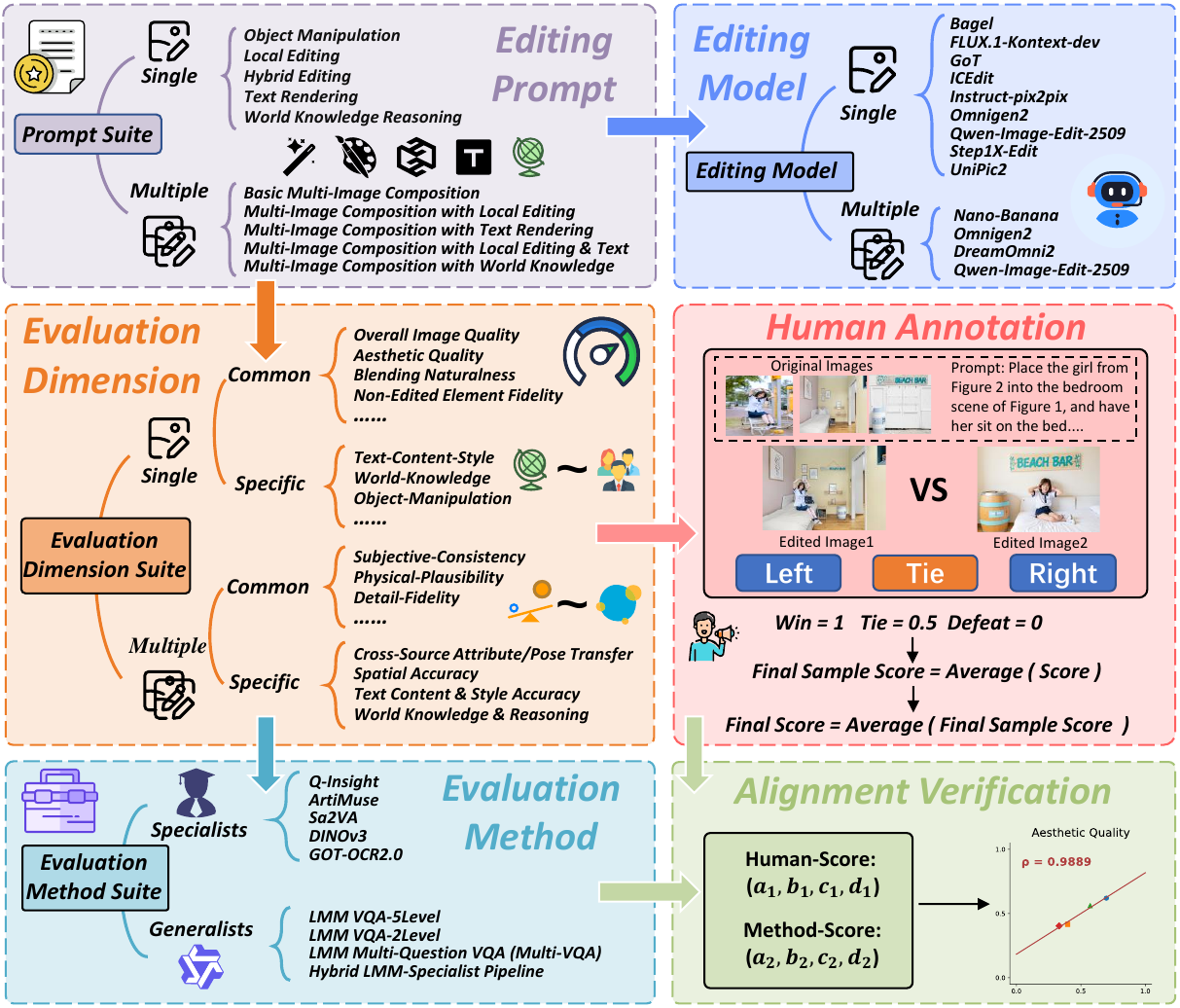}
    \vspace{-18pt}
    \caption{An overview of the proposed image-to-image editing evaluation benchmark suite, I2I-Bench. The process starts with our large-scale \textbf{Prompt Suite}, which defines the editing tasks. These prompts are fed into the \textbf{Editing Model} to edit images. The prompts also guide the selection of relevant dimensions from our hierarchical \textbf{Evaluation Dimension} Suite. Each dimension, in turn, specifies both the automated \textbf{Evaluation Method} Suite (combining Specialists and Generalists) and the criteria for \textbf{Human Annotation}. Finally, the results from the automated methods and human annotations are compared for \textbf{Alignment Verification} to ensure the reliability of our benchmark.}
    \vspace{-15pt}
    \label{fig:I2I-Bench}
\end{figure*}

To systematically address all the aforementioned limitations, we propose \textbf{I2I-Bench}, a comprehensive benchmark suite for evaluating image editing models. An overview of the entire benchmark suite is shown in Figure \ref{fig:I2I-Bench}. I2I-Bench aims to provide a comprehensive, automated, fine-grained, and human preference-aligned evaluation framework.
Firstly, our benchmark contains \textbf{diverse tasks} to facilitate comprehensive evaluation. Specifically, a total of 1000 prompts, systematically covering 10 carefully designed task categories across both from single-image and multi-image editing tsaks. Secondly, I2I-Bench contains comprehensive evaluation dimensions. We propose a hierarchical, decoupled evaluation framework with 30 fine-grained dimensions. For each dimension, we define a reproducible, automated hybrid evaluation pipeline, integrating specialized “specialist” tools for precision score prediction and general “generalists” for semantic understanding.
Thirdly, we conduct rigorous alignment validation for I2I-Bench. We conduct large-scale human preference experiments to validate our automated metrics. This validation confirms that our automated evaluation results, particularly those from our hybrid pipeline and LMM-based evaluators, achieve an extremely high consistency with human judgment, ensuring the reliability of our benchmark.
Finally, I2I-Bench provides \textbf{in-depth benchmarking and actionable insights}. Using this comprehensive suite, we evaluate numerous mainstream image editing models. This analysis reveals key performance trade-offs, identifies critical bottlenecks (such as failures in physical plausibility and abstract reasoning), and exposes universal limitations in current models. These findings provide actionable insights to guide future research toward addressing these fundamental gaps.

%% file: sec/2_related_work.tex
\section{Related Work}
\label{sec:related_work}

\subsection{Image Editing Models}
Image editing models have undergone rapid development in recent years. Early models such as Instruct-Pix2Pix \cite{brooks2023instructpix2pix} introduce instruction-based editing. Subsequent works such as Emu-Edit \cite{emuedit} and Qwen-Image-Edit \cite{wu2025qwenimagetechnicalreport} focus on improving editing fidelity and blending quality. Recently, the emergence of multi-reference editing (ME) models, represented by Nano-Banana \cite{nanobanana} and DreamOmni2 \cite{xia2025dreamomni2multimodalinstructionbasedediting}, have expanded the scope of possible edits to include more complex tasks such as cross-source attribute transfer and subject consistency. This rapid iteration of model capabilities underscores the growing need for more sophisticated evaluation methodologies \cite{huang2023vbench}.

\subsection{Image Editing Evaluation Benchmarks}
Existing evaluation benchmarks \cite{xu2025lmm4edit,editbench,tedbench,ma2024i2ebench,qian2025giebench,pan2025icebench,jia2025compbench} exhibit several limitations in their evaluation coverage of current model capabilities. For instance, TedBench \cite{tedbench} has a very limited test scope. EditBench \cite{editbench} covers multiple editing types but relies heavily on expensive and hard-to-reproduce manual annotations. To achieve automation, EditVal \cite{basu2023editval} adoptes LMMs \cite{vlm_survey} for evaluation but struggles with processing complex semantic and cognitive tasks. LMM4Edit \cite{xu2025lmm4edit} is also based on LMMs \cite{Qwen3VL}, but its evaluation dimensions remain restricted to a limited set of aspects such as perceptual quality, editing alignment, and attribute preservation. In summary, there remains a clear gap due to the lack of  a comprehensive benchmark capable of fine-grained, automated evaluation of advanced cognitive tasks, especially for emerging multi-image editing tasks.

%% file: sec/3_I2IBench_Suite.tex
\begin{figure*}[!t]
    \centering
    \includegraphics[width=\linewidth]{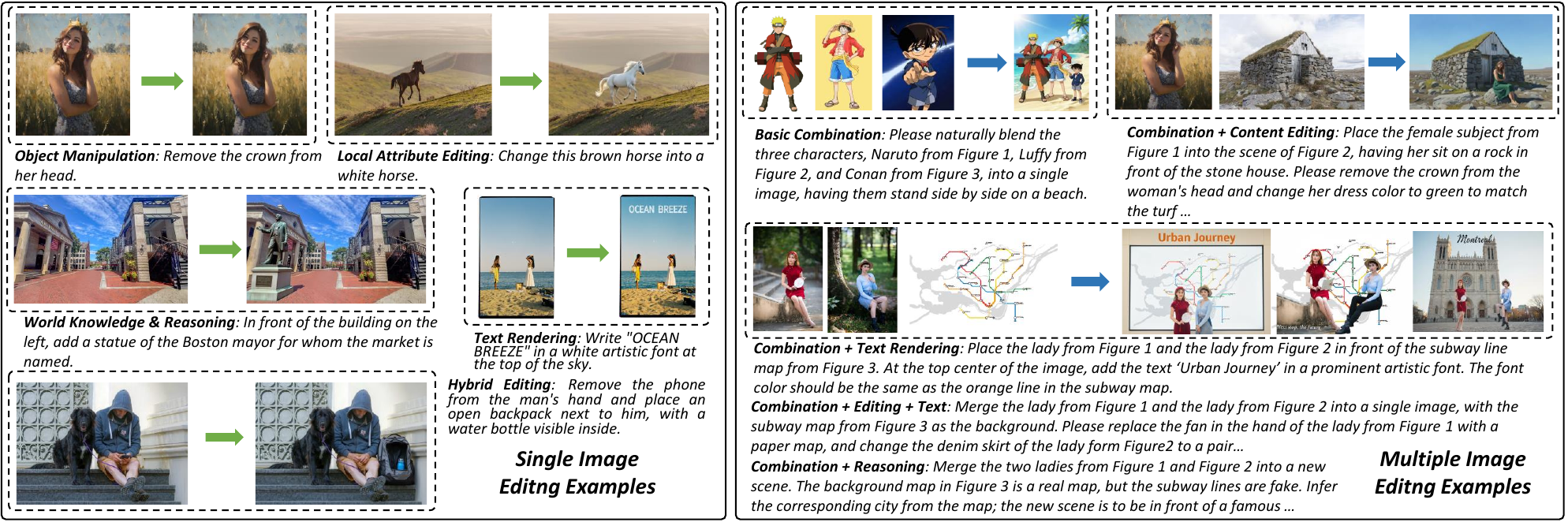}
    \vspace{-18pt}
    \caption{Visualization of the 10 task categories in the I2I-Bench Prompt Suite. The left half shows 5 single-image editing (SE) tasks, from “Object Manipulation” to “World Knowledge \& Reasoning”. The right half shows 5 multi-image editing (ME) tasks, illustrating increasing complexity from “Basic Combination” to “Combination + Reasoning”.}
    \label{fig:Edit-Sample}
    \vspace{-15pt}
\end{figure*}

\section{The I2I-Bench Suite}

I2I-Bench is a comprehensive benchmark suite composed of three integral components for the systematic evaluation of image editing models \cite{wu2025qwenimagetechnicalreport}: 1) a large-scale and structured Prompt Suite that provides diverse test cases; 2) a hierarchical Evaluation Dimension \& Method Suite that defines fine-grained aspects of editing quality with the corresponding quantitative methods; and 3) a Human Preference Annotation protocol designed to validate the alignment of our automated metrics with human judgment. An overview of the entire benchmark structure is shown in Figure~\ref{fig:I2I-Bench}.

\subsection{Prompt Suite}

The I2I-Bench Prompt Suite contains 1000 meticulously crafted prompts, equally split between single-image and multi-image editing tasks. This structured collection provides broad and systematic coverage of common editing scenarios and challenges. Illustrative examples for each prompt category are shown in Figure~\ref{fig:Edit-Sample}.

\noindent \textbf{Single-Image Editing Prompts.} This subset includes 500 prompts, structured as 100 base scenarios, each providing 5 prompts corresponding to the 5 SE categories, testing distinct aspects of single-image editing: \textit{Object Manipulation} (assessing fundamental global edits like adding, removing, or replacing objects); \textit{Local Editing} (evaluating precision in local attribute modification); \textit{Hybrid Editing} (challenging models with complex, multi-part instructions requiring both global and local changes); \textit{Text Rendering} (focusing on text accuracy); and \textit{World Knowledge \& Reasoning} (probing the application of external knowledge).

\noindent \textbf{Multi-Image Editing Prompts.} This subset contains 500 prompts, with 100 base scenarios covering 5 task categories of increasing complexity: \textit{Basic Combination} (testing simple subject extraction and composition); \textit{Combination + Content Editing} (requiring subsequent attribute/pose edits after combination); \textit{Combination + Text Rendering} (adding text to a composite scene); \textit{Combination + Editing + Text} (involving multi-step, mixed instructions); and \textit{Combination + Reasoning} (demanding integration of external knowledge based on the combined context).

\noindent \textbf{Relationship Between Prompts and Dimensions.} Our prompt categories (organized by user task scenarios) and evaluation dimensions (a fine-grained decomposition of quality) have a deliberate “many-to-many” design. This allows a single, complex prompt (\textit{e.g.,} from “Hybrid Editing”) to holistically assess multiple fine-grained quality aspects (\textit{e.g.,} “Blending Naturalness”, “Object Manipulation Accuracy”, and “Physical Plausibility”), simultaneously within an integrated task,  facilitating integrated quality assessment beyond isolated dimension testing.

\subsection{Evaluation Dimension and Method Suite}

To comprehensively evaluate the diverse and complex editing tasks defined in the Prompt Suite, we design a hierarchical Evaluation Dimension and Method Suite. Traditional image quality metrics (such as PSNR or LPIPS \cite{lpips}) only measure superficial pixel similarity and fail to assess the semantic accuracy, physical plausibility, or logical coherence in editing outputs. In contrast, our framework moves beyond low-level fidelity to evaluate deep semantic alignment in image editing.
Our evaluation suite comprises a total of 30 fine-grained evaluation dimensions, divided qually into 15 Single-Reference dimensions and 15 Multi-Reference dimensions. Within each category, the dimensions are further divided into “Common Dimensions”, applicable across all tasks for assessing fundamental quality, and “Specific Dimensions”, tailored to specific instruction types or editing scenarios.
Crucially, we define a reproducible evaluation pipeline for each dimension, integrating two categories of tools: “Specialists” and “Generalists”. Specialists refer to mature, quantitative tools trained for specific tasks (\textit{e.g.,} Q-Insight \cite{li2025qinsight}, ArtiMuse \cite{cao2025artimusefinegrainedimageaesthetics}, DINOv3 \cite{simeoni2025dinov3}). Generalists refer to powerful LMMs (Qwen3-VL-8B-Instruct \cite{Qwen3VL}), which we leverage to evaluate dimensions requiring complex semantic understanding, world knowledge, and reasoning.
In the following, we present the definitions of the dimensions integrated with their corresponding evaluation methods.

\subsubsection{Single-Reference Editing Dimensions and Methods}
The 15 single-reference dimensions are categorized into 7 Common Dimensions and 8 Specific Dimensions.

\subsubsection*{A. Common Dimensions}
Common dimensions assess the fundamental quality of all single-reference editing tasks.

\noindent \textbf{Overall Image Quality} and \textbf{Aesthetic Quality.} We evaluate the objective technical quality (\textit{e.g.,} clarity, noise) and the artistic appeal of the image, respectively. We use specialist models Q-Insight \cite{li2025qinsight} and ArtiMuse \cite{cao2025artimusefinegrainedimageaesthetics} to obtain direct quantitative scores for these two dimensions, respectively.
\par

\noindent \textbf{Blending Naturalness} and \textbf{Generative/Editing Artifacts.}  We evaluate the smoothness of the transition between the edited region and its surroundings, and the introduction of Artificial Intelligence (AI)-specific flaws (\textit{e.g.,} broken structures, unnatural textures). We employ the LMM VQA-5Level pipeline \cite{wu2023qalign,wang2025aigvassessor}, a method based on Visual Question Answering (VQA). The LMM \cite{Qwen3VL} is asked to choose from a 5-level scale \{“excellent”, “good”, “fair”, “poor”, “bad”\}, where each choice $c_i$ is assigned a weight $w_i \in \{1, 0.75, 0.5, 0.25, 0\}$. The LMM's logits $z_i$ are converted to probabilities $P_i$ via the Softmax function. The final score is the weighted average:
\begin{equation}
\label{eq:score_5level}
\text{Score}_{\text{5-level}}=\sum_{i=1}^{5}w_{i}\cdot P_{\text{LMM}}(c_{i}|I,Q),
\end{equation}
where $I$ represents the required image(s) for evaluation (\textit{e.g.,} $I_{\text{gen}}$, or the $I_{\text{orig}}, I_{\text{gen}}$ pair), and $Q$ is the question or instruction. For Editing Artifacts, the LMM receives $I_{\text{orig}}$, $I_{\text{gen}}$, and $Q$, while for Blending Naturalness, it focuses on the integration in $I_{\text{gen}}$.
\par

\noindent \textbf{Instruction Following (Macro).} We assess the model's overall understanding and execution of the instruction's core intent \cite{wang2025titscoreevaluatinglongpromptbased}. We reuse the LMM VQA-5level pipeline (Eq.~(\ref{eq:score_5level})) to evaluate the high-level semantic grasp.
\par

\noindent \textbf{Non-Edited Region Fidelity.} We assess whether image regions not targeted by the instruction remain unchanged post-edit. We use a hybrid LMM-specialist pipeline. This process combines LMM semantics with specialist precision: (1) The LMM receives the original/generated images and instruction, generating a segmentation command (\textit{e.g.,} “segment the horse”). (2) The specialist segmentation model Sa2VA \cite{yuan2025sa2va} uses this command to generate corresponding masks for the edited region, $M_{\text{edit}}$. (3) We invert this mask ($M_{\text{non-edit}} = \text{NOT}(M_{\text{edit}})$) to obtain the non-edited region mask. (4) The specialist feature extractor DINOv3 \cite{simeoni2025dinov3} extracts feature vectors $f_{\text{orig}}$ and $f_{\text{gen}}$ from within $M_{\text{non-edit}}$. (5) The final score is the cosine similarity between these vectors:
\begin{equation}
\label{eq:score_hybrid_sim}
\text{Score}_{\text{Hybrid-Sim}}=\frac{f_{\text{orig}}\cdot f_{\text{gen}}}{||f_{\text{orig}}||\cdot||f_{\text{gen}}||}.
\end{equation}

\par

\noindent \textbf{Physical Plausibility.} We evaluate whether the edit adheres to physical laws \cite{ps-diffusion} (\textit{e.g.,} lighting, perspective, gravity). We employ the LMM multi-question VQA (Multi-VQA) pipeline. This is a two-stage process: (1) Question Generation: The LMM receives the original image(s) and instruction, and is prompted to design N closed-ended (Yes/No) questions for the specific dimension (\textit{e.g.,} “Is the shadow direction of the new object correct?”). (2) Question Answering: The LMM receives the generated questions $Q_i$, original image(s), edited image, and instruction, and answers ‘Yes’ or ‘No’ to each. The final score is the ratio of ‘Yes’ answers:
\begin{equation}
\label{eq:score_multivqa}
\begin{gathered}
\text{Score}_{\text{Multi-VQA}} = \\
\frac{1}{N}\sum_{i=1}^{N} \mathbb{I}(\text{LMM}(I_{\text{gen}},I_{\text{orig}},Q_{i})=\text{‘Yes’}).
\end{gathered}
\end{equation}
\par

\subsubsection*{B. Specific Dimensions}
These 8 dimensions evaluate the execution quality of specific instructions.

\noindent \textbf{Object Manipulation Accuracy,} \textbf{Local Attribute Accuracy,} \textbf{Action/State Change Accuracy,} \textbf{Spatial Accuracy.} We evaluate task-specific accuracy, such as the correctness of “add/remove/replace” object operations, the accurate application of  “change color/material” attribute modifications, and the faithful execution of spatial descriptions (\textit{e.g.,} “to the left of”). We primarily use the LMM VQA-2level pipeline \cite{vqascore}. The LMM is asked a binary (Yes/No) question regarding the success of the specific editing operation (\textit{e.g.,} “Has the horse's color been successfully changed to white?”). The final score is the probability of the 'Yes' answer:
\begin{equation}
\label{eq:score_2level}
\text{Score}_{\text{2-level}} = P_{\text{LMM}} (\text{‘Yes’} | I, Q).
\end{equation}

\par

\noindent \textbf{Text Content \& Style Accuracy.} We assess the model's ability to render text in images, with correct spelling and appropriate visual presentation. We use a multi-step hybrid pipeline: (1) Content Accuracy: We use the specialist OCR model GOT-OCR2.0 \cite{got-ocr} to compute $S_c$ based on its distance (\textit{e.g.,} Levenshtein) from the target text. (2) Style and Position: We use the LMM VQA-5level pipeline (Eq.~(\ref{eq:score_5level})), instructing the LMM to ignore spelling and provide a 1-5 discrete score ($S_s$) based on style and position match. (3) Score Fusion: The final score ($\text{Score}_{\text{final}}$) is calculated by combining $S_s$ (style score) and $S_c$ (content accuracy) as follows:
\begin{equation}
\label{eq:score_final_text}
\text{Score}_{\text{final}} = \left( \frac{S_s - 1}{4} \right) \times
\begin{cases} 
      1.0 & \text{if } S_c = 1.0 \\
      0.8 & \text{if } 0.8 \le S_c < 1.0 \\
      0.5 & \text{if } 0.6 \le S_c < 0.8 \\
      0.1 & \text{if } S_c < 0.6  .
\end{cases}
\end{equation}
\par

\noindent \textbf{World Knowledge \& Reasoning.} We evaluate instructions requiring external knowledge (\textit{e.g.,} “add a statue of ...”). We use the LMM VQA-2level pipeline \cite{vqascore} (Eq.~(\ref{eq:score_2level})) to judge if the generated image matches a pre-defined correct answer set.
\par

\noindent \textbf{Subject Identity Fidelity.} We assesses whether a subject's core identity is preserved during attribute modifications. We still use the hybrid LMM-specialist pipeline (Eq.~(\ref{eq:score_hybrid_sim})) to compute feature similarity between the original and edited images, but we only compare the segmented subject regions directly (without mask inversion).
\par

\noindent \textbf{Composition \& Interaction.} We evaluate the logical interaction and visual coherence between newly introduced or altered elements and existing scene content. We reuse the LMM multi-question VQA pipeline (Eq.~(\ref{eq:score_multivqa})) to ask specific questions about interaction plausibility.
\par
\begin{table*}[!t]
  \centering
  \caption{Results of the single-image editing benchmark. All scores are normalized; higher is better. Best scores are highlighted.}
  \vspace{-8pt}
  \label{tab:se_bench_scaled}
  
  \resizebox{0.9\textwidth}{!}{%
    \begin{tabular}{l c c c c c c c c}
    
    \toprule
    \textbf{Model} &
    \makecell[c]{Action-\\State-Change} &
    \makecell[c]{Aesthetic-\\Quality} &
    \makecell[c]{Blending-\\Naturalness} &
    \makecell[c]{Composition-\\Interaction} &
    \makecell[c]{Editing-\\Artifacts} &
    \makecell[c]{Image-\\Quality} &
    \makecell[c]{Instruction-\\Following} &
    \makecell[c]{Local-\\Attribute} \\
    \midrule
    Qwen-Image-Edit-2509 \cite{wu2025qwenimagetechnicalreport} & \best{0.977} & 0.589 & \best{0.807} & \best{0.823} & \best{0.907} & 0.811 & \best{0.930} & \best{0.961} \\
    Step1X-Edit \cite{liu2025step1x-edit}       & 0.819 & 0.584 & 0.738 & 0.804 & 0.838 & 0.807 & 0.850 & 0.895 \\
    UniPic-2 \cite{wei2025skyworkunipic20building}         & 0.964 & 0.567 & 0.733 & 0.819 & 0.798 & 0.799 & 0.862 & 0.941 \\
    Bagel  \cite{deng2025bagel}    & 0.781 & 0.584 & 0.669 & 0.752 & 0.756 & 0.810 & 0.785 & 0.865 \\
    FLUX.1-Kontext-dev \cite{labs2025flux1kontextflowmatching} & 0.882 & \best{0.592} & 0.697 & 0.753 & 0.753 & \best{0.823} & 0.787 & 0.851 \\
    Omnigen2 \cite{wu2025omnigen2} & 0.844 & 0.569 & 0.602 & 0.750 & 0.694 & 0.812 & 0.755 & 0.848 \\
    ICEdit \cite{icedit} & 0.722 & 0.590 & 0.599 & 0.734 & 0.623 & 0.797 & 0.680 & 0.784 \\
    GoT \cite{gotedit} & 0.659 & 0.563 & 0.431 & 0.621 & 0.444 & 0.786 & 0.490 & 0.606 \\
    instruct-pix2pix \cite{brooks2023instructpix2pix} & 0.343 & 0.552 & 0.176 & 0.542 & 0.173 & 0.769 & 0.221 & 0.404 \\
    
    \midrule
    
    \textbf{Model} &
    \makecell[c]{Non-Edited-\\Fidelity} &
    \makecell[c]{Object-\\Manipulation} &
    \makecell[c]{Physical-\\Plausibility} &
    \makecell[c]{Spatial-\\Accuracy} &
    \makecell[c]{Subject-\\Identity} &
    \makecell[c]{Text-Content-\\Style} &
    \makecell[c]{World-\\Knowledge} &
    \textbf{Overall} \\
    \midrule
    Qwen-Image-Edit-2509 \cite{wu2025qwenimagetechnicalreport} & 0.898 & \best{0.937} & 0.527 & \best{0.951} & 0.768 & \best{0.666} & \best{0.749} & \best{0.813} \\
    Step1X-Edit \cite{liu2025step1x-edit}       & 0.928 & 0.862 & 0.537 & 0.844 & 0.799 & 0.456 & 0.655 & 0.773 \\
    UniPic-2 \cite{wei2025skyworkunipic20building}         & 0.818 & 0.927 & \best{0.562} & 0.930 & 0.677 & 0.372 & 0.605 & 0.767 \\
    Bagel  \cite{deng2025bagel}    & \best{0.938} & 0.838 & 0.502 & 0.808 & \best{0.830} & 0.438 & 0.668 & 0.742 \\
    FLUX.1-Kontext-dev \cite{labs2025flux1kontextflowmatching} & 0.852 & 0.804 & 0.479 & 0.813 & 0.694 & 0.404 & 0.570 & 0.727 \\
    Omnigen2 \cite{wu2025omnigen2} & 0.891 & 0.810 & 0.488 & 0.793 & 0.764 & 0.434 & 0.532 & 0.713 \\
    ICEdit \cite{icedit} & 0.902 & 0.715 & 0.518 & 0.732 & 0.781 & 0.265 & 0.446 & 0.684 \\
    GoT \cite{gotedit} & 0.897 & 0.596 & 0.399 & 0.500 & 0.760 & 0.038 & 0.469 & 0.575 \\
    instruct-pix2pix \cite{brooks2023instructpix2pix} & 0.742 & 0.285 & 0.348 & 0.247 & 0.647 & 0.014 & 0.300 & 0.416 \\
    \bottomrule
    
    \end{tabular}%
  } 
  
  \vspace{-10pt}
\end{table*}
\begin{table*}[!t]
  \centering
  \caption{Results of the multi-image editing benchmark. All scores are normalized; higher is better. Best scores are highlighted.}
  \vspace{-8pt}
  \label{tab:me_bench_scaled}
  
  \resizebox{0.9\textwidth}{!}{%
    \begin{tabular}{l c c c c c c c c}
    
    \toprule
    \textbf{Model} &
    \makecell[c]{Aesthetic-\\Quality} &
    \makecell[c]{Blending-\\Naturalness} &
    \makecell[c]{Composition-\\Interaction} &
    \makecell[c]{Cross-Source-\\Attribute} &
    \makecell[c]{Detail-\\Fidelity} &
    \makecell[c]{Image-\\Quality} &
    \makecell[c]{Instruction-\\Following} &
    \makecell[c]{Inter-Subject-\\Consistency} \\
    \midrule
    Nano-Banana \cite{nanobanana}       & \best{0.580} & \best{0.562} & \best{0.693} & \best{0.700} & \best{0.514} & 0.803 & \best{0.830} & \best{0.582} \\
    Qwen-Image-Edit-2509 \cite{wu2025qwenimagetechnicalreport} & 0.566 & 0.556 & 0.665 & 0.628 & 0.506 & 0.800 & 0.744 & 0.564 \\
    DreamOmni2 \cite{xia2025dreamomni2multimodalinstructionbasedediting} & 0.555 & 0.579 & 0.604 & 0.510 & 0.478 & \best{0.827} & 0.658 & 0.556 \\
    Omnigen2 \cite{wu2025omnigen2} & 0.531 & 0.546 & 0.587 & 0.518 & 0.462 & \best{0.827} & 0.620 & 0.543 \\
    
    \midrule
    
    \textbf{Model} &
    \makecell[c]{Non-Edited-\\Fidelity} &
    \makecell[c]{Physical-\\Plausibility} &
    \makecell[c]{Spatial-\\Accuracy} &
    \makecell[c]{Subject-\\Consistency} &
    \makecell[c]{Subject-\\Extraction} &
    \makecell[c]{Text-Content-\\Style} &
    \makecell[c]{World-\\Knowledge} &
    \textbf{Overall} \\
    \midrule
    Nano-Banana \cite{nanobanana}       & \best{0.655} & \best{0.433} & \best{0.894} & \best{0.535} & \best{0.530} & \best{0.687} & \best{0.721} & \best{0.636} \\
    Qwen-Image-Edit-2509 \cite{wu2025qwenimagetechnicalreport} & 0.613 & 0.424 & 0.893 & 0.516 & 0.505 & 0.660 & 0.044 & 0.604 \\
    DreamOmni2 \cite{xia2025dreamomni2multimodalinstructionbasedediting} & 0.609 & 0.388 & 0.762 & 0.472 & 0.434 & 0.323 & 0.210 & 0.562 \\
    Omnigen2 \cite{wu2025omnigen2} & 0.564 & 0.346 & 0.749 & 0.463 & 0.445 & 0.479 & 0.038 & 0.545 \\
    \bottomrule
    
    \end{tabular}%
  } 
  
  \vspace{-10pt}
\end{table*}

\subsubsection{Multi-Reference Editing Dimensions and Methods}
The 15 multi-reference dimensions consist of 9 shared dimensions (reused from single-reference, with identical evaluation methods) and 6 multi-reference-specific dimensions. These specific dimensions focus on evaluating cross-image composition and consistency.

\noindent \textbf{Non-Edited Element Fidelity.} Assesses whether elements from source images that are not designated as editing targets (\textit{e.g.,} background elements) are erroneously modified or discarded during composition. We reuse the Hybrid LMM-Specialist pipeline (Eq.~\ref{eq:score_hybrid_sim}). The LMM segments elements to be preserved, inverts the corresponding mask, then DINOv3 \cite{simeoni2025dinov3} is applied to compute similarity on the non-edited elements.
\par

\noindent \textbf{Subject Consistency} and \textbf{Detail Fidelity/Preservation.} We assess the preservation of a subject's identity and fine details from its source image to the newly generated image. We reuse the hybrid LMM-specialist pipeline (Eq.~(\ref{eq:score_hybrid_sim})) to compute feature similarity, by directly comparing the segmented subjects from the source and edited images (without mask inversion).
\par

\noindent \textbf{Subject Extraction \& Composition.} We evaluate whether subjects are completely extracted and correctly composed into the new scene. We implement a two-step multiplicative scoring process: (1) The LMM VQA-2level pipeline \cite{vqascore} (Eq.~(\ref{eq:score_2level})) provides a binary score ($\text{Score}_{\text{count}}$) for the correct number of subjects. (2) Then multiplied by the $\text{Score}_{\text{consistency}}$ (calculated via the \textbf{Subject Consistency} hybrid pipeline) to yield $\text{Score}_{\text{final}} = \text{Score}_{\text{count}} \times \text{Score}_{\text{consistency}}$.
\par

\noindent \textbf{Inter-Subject Consistency} and \textbf{Cross-Source Attribute/Pose Transfer.} We assess the visual coherence of subjects from different sources when composed together (e.g., in lighting, scale, and style); or evaluate the fidelity of transferring attributes or poses between subjects across sources. We still use the LMM VQA-2level (Eq.~(\ref{eq:score_2level})) or Multi-VQA (Eq.~(\ref{eq:score_multivqa})) pipelines to pose specific questions about these complex cross-image interactions.
\par

\subsection{Human Preference Annotation}

To validate the alignment of our proposed automated evaluation dimensions (especially those relying on LMMs) with genuine human perception, we conduct a large-scale human preference annotation experiment, following established practices in benchmark development.

\noindent\textbf{Model and Sample Selection.} We evaluate 9 single image editing and 4 multi-image editing models. For ME tasks, all 4 models are compared in a pairwise manner. For SE tasks, to manage annotation costs, we randomly assign a unique combination of 4 models for comparison within each of the 15 SE dimensions. We uniformly sample 85 items (510 pairs) for all 30 evaluation dimensions to ensure broad and representative coverage.

\noindent\textbf{Annotation Process.} We use a pairwise comparison (A vs. B vs. Tie) format. For each sample, annotators are provided with the source image(s), instruction, and a clear definition of the specific dimension being evaluated. They are strictly instructed to judge only on that single dimension, disregarding all other aspects. For example, when evaluating Subject Identity Fidelity, annotators are instructed to select the image that better preserves subject identity, even if it exhibites noticeable artifacts in Blending Naturalness artifacts.

\noindent\textbf{Win Ratio Calculation.} We calculate a Win Ratio for each model per dimension based on pairwise results. In this calculation, a 'win' is assigned a numerical value of 1, a 'tie' is assigned 0.5, and a 'loss' is assigned 0. The final Win Ratio for each model in a specific dimension is calculated as the total score accumulated divided by the number of comparisons it participated for that dimension.

%% file: sec/4_Experiment.tex
\section{Experiments}
\label{sec:experiments}

In this section, we conduct a comprehensive evaluation of a series of advanced image editing models using the proposed I2I-Bench.
We first present the detailed performance of all models across 30 evaluation dimensions in Subsection \ref{subsec:eval_by_dimension}.
We then validate the consistency between the I2I-Bench evaluation methodology and human perception through large-scale human preference annotation.
Finally, we provide a deeper analysis by task category in Subsection \ref{subsec:eval_by_category} and a comparative analysis between single-image and multi-image editing models in Subsection \ref{subsec:se_vs_me}.

\begin{table*}[!t]
  \centering
  \caption{\textbf{Human Preference Alignment and Ablation Study.} This table presents the Pearson correlation ($\rho$) coefficients between I2I-Bench automated metrics and human preferences (Win Ratio). The results show extremely high consistency across all 30 dimensions, strongly validating the reliability of our evaluation methodology. The table also presents an ablation study comparing our hybrid \textbf{I2I-Bench} pipeline vs. a \textbf{Pure LMM-Baseline} (general LMM VQA-5level), which validates our hybrid design.}
  \label{tab:ablation_and_alignment_transposed_8x8}
  
  \resizebox{0.8\textwidth}{!}{%
  
    \begin{tabular}{l c c c c c c c c} 
    
      \toprule
      \textbf{Metric (SE)} &
      \makecell[c]{Action-\\State-Change} &
      \makecell[c]{Aesthetic-\\Quality} &
      \makecell[c]{Blending-\\Naturalness} &
      \makecell[c]{Composition-\\Interaction} &
      \makecell[c]{Editing-\\Artifacts} &
      \makecell[c]{Image-\\Quality} &
      \makecell[c]{Instruction-\\Following} &
      \makecell[c]{Local-\\Attribute} \\
      \midrule
      \textbf{Ours ($\rho$)} & \textbf{0.9839} & \textbf{0.9889} & \textbf{0.8866} & \textbf{0.9997} & \textbf{0.9006} & \textbf{0.9033} & \textbf{0.9960} & \textbf{0.9877} \\
      LMM-Baseline ($\rho$) & 0.9769 & -0.0643 & 0.8866 & 0.9851 & 0.8983 & 0.4154 & 0.9960 & 0.9842 \\
      \midrule 
      
      \textbf{Metric} &
      \makecell[c]{Non-Edited-\\Fidelity} &
      \makecell[c]{Object-\\Manipulation} &
      \makecell[c]{Physical-\\Plausibility} &
      \makecell[c]{Spatial-\\Accuracy} &
      \makecell[c]{Subject-\\Identity} &
      \makecell[c]{Text-Content-\\Style} &
      \makecell[c]{World-\\Knowledge} &
      \textbf{Overall} \\
      \midrule
      \textbf{Ours ($\rho$)} & \textbf{0.9019} & \textbf{0.9787} & \textbf{0.8055} & \textbf{0.9303} & \textbf{0.9133} & \textbf{0.9979} & \textbf{0.9628} & \textbf{0.9425} \\
      LMM-Baseline ($\rho$) & 0.7687 & 0.9673 & 0.6958 & 0.9191 & -0.3494 & 0.8847 & 0.9506 & 0.7277 \\
      \midrule 

      \textbf{Metric (ME)} &
      \makecell[c]{Aesthetic-\\Quality} &
      \makecell[c]{Blending-\\Naturalness} &
      \makecell[c]{Composition-\\Interaction} &
      \makecell[c]{Cross-Source-\\Attribute} &
      \makecell[c]{Detail-\\Fidelity} &
      \makecell[c]{Image-\\Quality} &
      \makecell[c]{Instruction-\\Following} &
      \makecell[c]{Inter-Subject-\\Consistency} \\
      \midrule
      \textbf{Ours ($\rho$)} & \textbf{0.9034} & \textbf{0.6177} & \textbf{0.8951} & \textbf{0.9917} & \textbf{0.8767} & \textbf{0.6579} & \textbf{0.9248} & \textbf{0.9469} \\
      LMM-Baseline ($\rho$) & 0.7138 & 0.6177 & 0.8205 & 0.9902 & 0.8092 & -0.9202 & 0.9248 & 0.8293 \\
      \midrule 
      
      \textbf{Metric} &
      \makecell[c]{Non-Edited-\\Fidelity} &
      \makecell[c]{Physical-\\Plausibility} &
      \makecell[c]{Spatial-\\Accuracy} &
      \makecell[c]{Subject-\\Consistency} &
      \makecell[c]{Subject-\\Extraction} &
      \makecell[c]{Text-Content-\\Style} &
      \makecell[c]{World-\\Knowledge} &
      \textbf{Overall} \\
      \midrule
      \textbf{Ours ($\rho$)} & \textbf{0.6459} & \textbf{0.9017} & \textbf{0.9008} & \textbf{0.9401} & \textbf{0.9732} & \textbf{0.9043} & \textbf{0.9523} & \textbf{0.8688} \\
      LMM-Baseline ($\rho$) & 0.5978 & 0.8887 & 0.8230 & 0.6616 & 0.7763 & 0.8573 & 0.8534 & 0.6829 \\
      \bottomrule 
      
    \end{tabular}%
  } 
  \vspace{-3pt}
\end{table*}

\begin{figure*}[!t]
    \centering
    \includegraphics[width=1\linewidth]{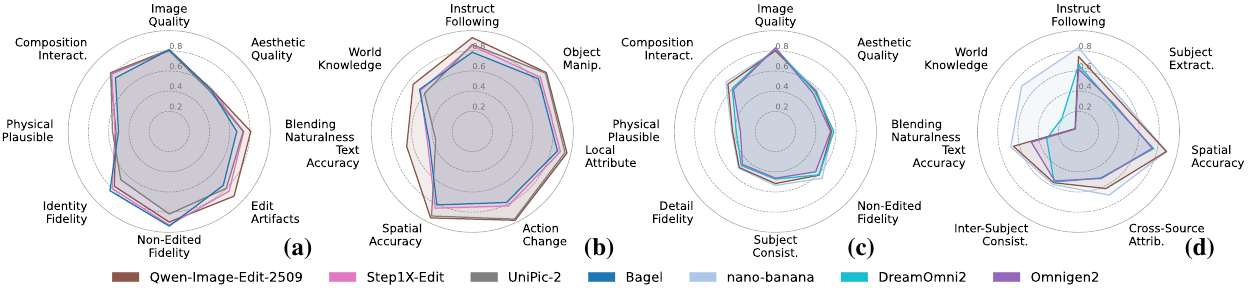}
    \caption{Capability radar charts for the evaluated models on key dimensions. (a) Foundational Quality \& Fidelity (SE models). (b) Task Execution \& Advanced Capabilities (SE models). (c) Foundational Quality \& Fidelity (ME models). (d) Task Execution \& Advanced Capabilities (ME models).}
    \label{fig:leida}
    \vspace{-2pt}
\end{figure*}

\subsection{Evaluation by Dimension}
\label{subsec:eval_by_dimension}

We first present the main evaluation results of I2I-Bench across all 30 fine-grained dimensions. The detailed scores for all nine SE models and four ME models are presented in Table \ref{tab:se_bench_scaled} and Table \ref{tab:me_bench_scaled}, respectively. Key performance trends and trade-offs are visualized in Figure \ref{fig:leida}.

For SE models, as illustrated in Figure \ref{fig:leida}(a)-(b),  Qwen-Image-Edit-2509 \cite{wu2025qwenimagetechnicalreport} excels in “Blending Naturalness” and “Editing Artifacts”, and leads in most task execution dimensions (\textit{e.g.,} “Instruction-Following-Macro”). We also identify clear model-specific trade-offs: Bagel \cite{deng2025bagel}, for instance, attains high score on “Non-Edited-Element-Fidelity” but low on “Blending Naturalness”, highlighting a conflict between preserving the background and blending new content.

For ME models, Figure \ref{fig:leida}(c-d) reveals that these tasks pose substantially greater challenges, with foundational quality scores (\textit{e.g.,} “Aesthetic Quality”) being generally lower than those in SE tasks. Nano-banana \cite{nanobanana} demonstrates strong overall performance, particularly in “World-Knowledge-Reasoning”, where other models like Omnigen2 \cite{wu2025omnigen2} and Qwen-Image-Edit-2509 \cite{wu2025qwenimagetechnicalreport} completely fail. This significant disparity, which highlights a fundamental gap in reasoning capabilities, is analyzed in detail in Section \ref{subsec:se_vs_me}.

\subsection{Validation of Human Alignment}
\label{subsec:human_alignment}

\begin{figure*}[!t]
    \centering
    \includegraphics[width=\linewidth]{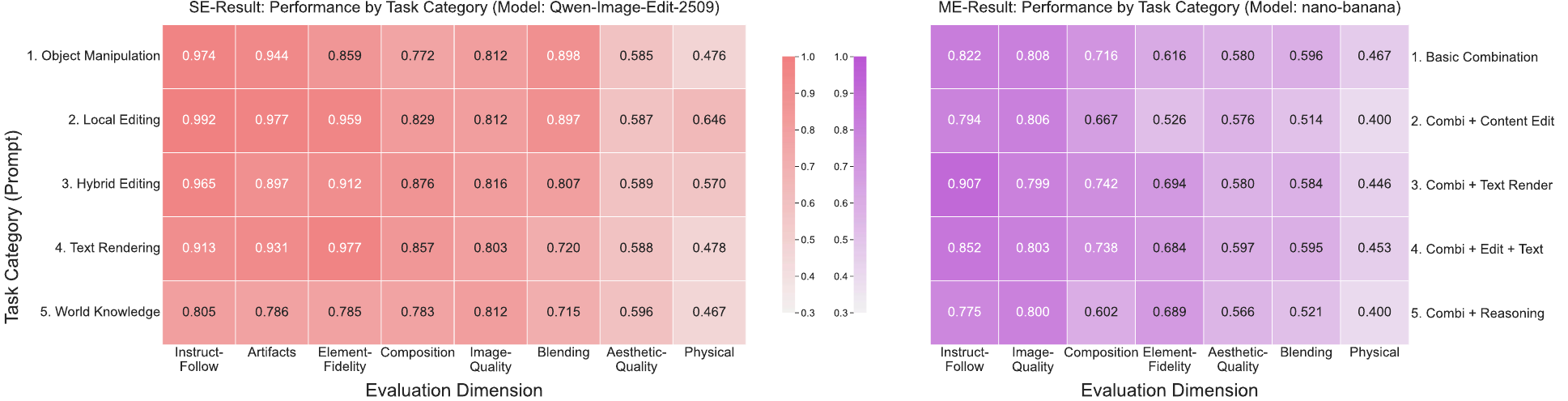}
    \vspace{-16pt}
    \caption{Performances of top-performing SE and ME models on common dimensions across task categories. (1) The performance of Qwen-Image-Edit-2509 (SE) as task cognitive complexity increases. (2) The performance of nano-banana (ME) varies across complex combination tasks.}
    \label{fig:reli}
    \vspace{-18pt}
\end{figure*}

The credibility of an evaluation benchmark critically depends on its alignment with human perception. To verify the reliability of I2I-Bench (especially our LMM-based evaluators), we conduct a large-scale human preference annotation experiment, as described in Subsection \ref{subsec:human_alignment}. We adopt a pairwise comparison format and calculated a human preference “Win Ratio” for each model in each dimension. We then compute the Pearson correlation coefficient ($\rho$) between the I2I-Bench automated evaluation scores and the human preference win ratios. The results are presented in Table \ref{tab:ablation_and_alignment_transposed_8x8} (see the “Ours ($\rho$)” rows). We observe a high consistency between our automated results and human judgment. As shown in the table, across all 30 dimensions, the correlation coefficients are excellent. This result validates the effectiveness of I2I-Bench as a reliable benchmark.

\subsection{Ablation Study on Evaluation Pipeline}
To validate the necessity of our hybrid evaluation paradigm, we conduct an ablation study comparing our method against a “Pure LMM” baseline that uses the LMM VQA-5level as the general-purpose evaluation method. As presented in Table~\ref{tab:ablation_and_alignment_transposed_8x8}, a direct comparison between the “Ours ($\rho$)” and “LMM-Baseline ($\rho$)” rows strongly supports the design of I2I-Bench: the LMM baseline fails catastrophically on dimensions requiring specialist perception (\textit{e.g.}, Aesthetic-Quality $\rho = -0.0643$ vs. our $0.9889$), proving our “specialist” tools are indispensable. Simultaneously, our designed “hybrid LMM-specialist” pipeline also significantly outperforms the LMM baseline on complex compositional tasks. For specific accuracy dimensions, our selected LMM VQA-2level method offers superior interpretability---its probability output ($P(\text{‘Yes’}$)) allows for an intuitive 0.5 threshold to classify an operation as “successful,” a property the 5-level score lacks. This ablation study confirms that I2I-Bench's hybrid design is critical for achieving robust, interpretable, and highly human-aligned evaluation.

\subsection{Evaluation by Task Category}
\label{subsec:eval_by_category}

The I2I-Bench Prompt Suite is systematically organized into distinct task categories. To identify model-specific limitations, we analyze the average performance on common dimensions across these categories. As illustrated in Figure \ref{fig:reli}, for the top-performing SE model Qwen-Image-Edit-2509 \cite{wu2025qwenimagetechnicalreport}, foundational quality (“Image-Quality”) remains stable, but the cognitive complexity of the task significantly impacts key capabilities. This trend is most evident in the “Instruction-Following-Macro” dimension, showing a steady decline from “Local Editing” (0.992) to “World Knowledge” (0.805). A similar decrease is observed in “Blending-Naturalness” (from 0.897 to 0.715), indicating that as the task's cognitive load increases, the model's ability to follow instructions and maintain fidelity degrades.

For the top-performing ME model nano-banana, as illustrated in Figure \ref{fig:reli}, we again observe stable “Image-Quality” but identify two distinct bottlenecks.
First, a “semantic bottleneck” emerges in the “Combi + Content Edit” task, the model achieves its lowest scores in fidelity-related dimensions such as “Blending-Naturalness” (0.514) and “Physical-Plausibility” (0.400). This suggests the complex semantic task of “combining and then deeply modifying” is a key challenge.
Second, a “logic bottleneck” is evident in the “Combi + Reasoning” task, the model performs worst in “Composition-Interaction” (0.602) and “Instruction-Following-Macro” (0.775), indicating that abstract logic requirements hinder instruction understanding.

\subsection{Comparative Analysis: SE vs. ME}
\label{subsec:se_vs_me}

\begin{figure}
    \centering
    \includegraphics[width=1\linewidth]{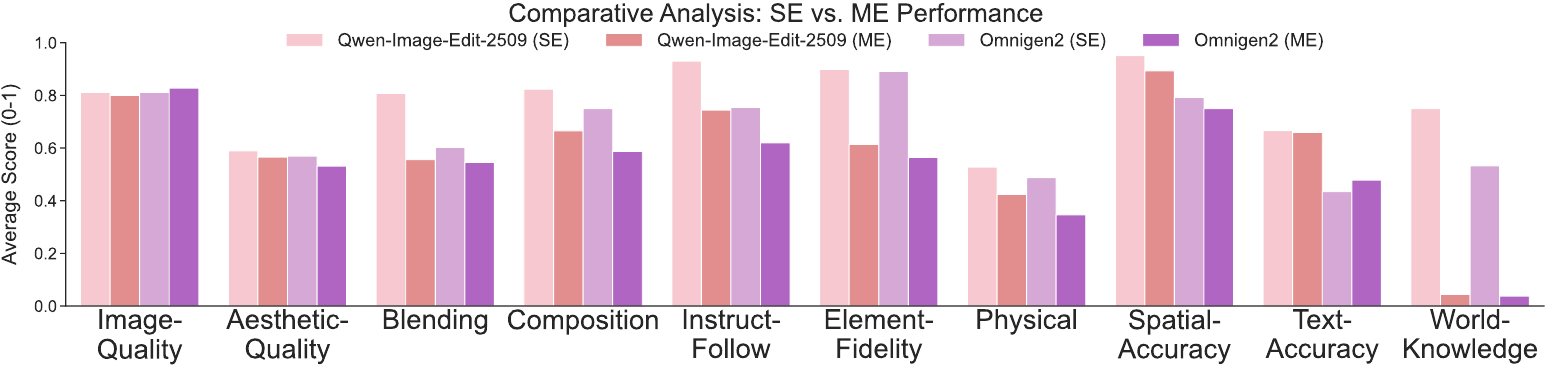}
    \vspace{-15pt}
    \caption{Performance comparison between Single-Image Editing (SE) and Multi-Image Editng (ME) tasks for Qwen-Image-Edit-2509 and Omnigen2 on shared dimensions.}
    \label{fig:duibi}
    \vspace{-13pt}
\end{figure}

Qwen-Image-Edit-2509 \cite{wu2025qwenimagetechnicalreport} and Omnigen2 \cite{wu2025omnigen2} support both SE and ME tasks, allowing for a direct comparison of the challenges posed by multi-image inputs. As shown in Figure \ref{fig:duibi}, both models exhibit a general performance degradation when shifting from SE to ME tasks. For Qwen-Image-Edit-2509, its “Blending-Naturalness” score drops from 0.807 (SE) to 0.556 (ME), and “Instruction-Following-Macro” drops from 0.930 (SE) to 0.744 (ME). These results indicate that processing and fusing information from multiple image sources impose significantly higher demands on model capabilities.

However, we must distinguish between performance drops on comparable metrics and fundamental increases in task difficulty.
First, some metrics have different nuances (\textit{e.g.,} “Non-Edited Fidelity” in SE refers to “Region” fidelity, whereas in ME it refers to “Element” fidelity).
Second, the most dramatic difference is in the “World-Knowledge-Reasoning” dimension, where both models' scores plummet from acceptable SE levels (Qwen-Image-Edit-2509: 0.749, Omnigen2: 0.533) to near-zero (Qwen-Image-Edit-2509: 0.044, Omnigen2: 0.038). This is not a “capability loss” but reflects our intentional benchmark design. To test the upper limits of advanced models, our ME prompts for this dimension involve high-difficulty abstract reasoning (\textit{e.g.,} Sudoku, map recognition) far more complex than the knowledge retrieval tasks in SE. This result demonstrates that the “knowledge retrieval” capability shown in SE tasks does not extend to the “abstract logical reasoning” required in our ME tasks, revealing a significant capability gap.

%% file: sec/6_conclusion.tex
\section{Conclusion}
We propose I2I-Bench, a comprehensive benchmark for image-to-image editing \cite{wu2025qwenimagetechnicalreport,nanobanana}, featuring a systematic prompt suite (10 categories) and a novel hybrid “Specialist-Generalist” evaluation system (30 dimensions). Our large-scale alignment study validates I2I-Bench's high consistency with human judgment and the necessity of our hybrid design. Using this benchmark, we reveal key cognitive trade-offs and expose several critical universal limitations in current models, particularly their failures in Physical Plausibility, multi-lingual Text Rendering, and Abstract Logical Reasoning. We believe I2I-Bench provides a valuable tool to guide future research toward addressing these fundamental gaps.


%% file: sec/Appendix.tex
\clearpage          
\appendix           
\section*{Appendix} 

\section{Model Details}

\subsection{Evaluated Models}

We evaluated 9 Single-Image (SE) editing models and 4 Multi-Image (ME) editing models, as shown in the main paper.
For nano-banana, we obtained evaluation results through its official API interface. For all other models (including Qwen-Image-Edit-2509, Bagel, Step1X-Edit, UniPic-2, FLUX.1-Kontext-dev, Omnigen2, ICEdit, GoT, instruct-pix2pix, and DreamOmni2), we used their publicly available checkpoints. During inference, we uniformly adopted the default inference configurations provided in the models' official repositories or \texttt{diffusers} library. No additional parameter tuning was performevd, ensuring a fair and standardized evaluation.

\subsection{Evaluation Tools}

Our evaluation pipeline utilizes both generalist and specialist models. For the Generalist Evaluator, all LMM-based evaluation pipelines detailed in Appendix F (LMM VQA-5Level, LMM VQA-2Level, and LMM Multi-Question VQA) were conducted using Qwen3-VL-8B-Instruct. For the Specialist Tools, we utilized Q-Insight, ArtiMuse, GOT-OCR2.0, Sa2VA, and DINOv3, all employed with their default public settings.

\section{Human Preference Annotation Details}

To rigorously validate the alignment of I2I-Bench's automated evaluation methods (especially the LMM evaluators) with genuine human perception, we conducted a large-scale human preference annotation experiment.

\textbf{Annotator Guidelines.} We employed a pairwise comparison format. Annotators were shown images generated by two different models (\textit{e.g.}, Model A and Model B) for the same prompt and dimension, and were asked to select “A is better,” “B is better,” or “Tie.” The most critical principle was that annotators were strictly instructed to judge \textit{solely based on the single dimension being evaluated}, and to disregard all other quality aspects. For example, when evaluating “Subject Identity Fidelity,” annotators were told: “You must choose the image that better preserves the subject's identity (\textit{e.g.}, face, features). Even if the other image has better blending or fewer artifacts, you must penalize it if the subject's identity is distorted." Conversely, for “Blending Naturalness,” the guidance was: “You must focus only on whether the transition of the edited region is smooth and seamless. Even if the image did not follow the instruction perfectly, you should choose it if its blending is superior.” We provided detailed manuals with positive and negative examples for all 30 fine-grained dimensions to ensure a consistent understanding among all annotators.

\textbf{Quality Assurance.} To ensure the accuracy and consistency of the annotated data, we implemented a rigorous, multi-step quality assurance process. First, we (the authors) prepared clear definitions, criteria, and “what to look for” vs. “what to ignore” examples for all 30 dimensions. Second, before the main task, all annotators had to complete a “Pre-Labeling Trial” of approximately 30 pairwise comparison samples. Third, we reviewed these trial results and provided one-on-one feedback to annotators to clarify any misunderstandings and unify the standards. Fourth, we iterated on the guidelines, supplementing them with confusing cases found during the trial. Finally, after all annotations were complete, we (the authors) randomly sampled 20\% of the total annotations from each dimension for post-labeling checks. If the error rate (disagreement with the authors) in this sample exceeded 10\%, all data for that dimension was considered invalid and re-assigned to a different annotator for re-labeling. This strict training and QA process ensures our human preference data is highly reliable for alignment validation.

\section{Rationale for Evaluation Methods}
\label{sec:appendix_rationale}

In I2I-Bench, we firmly contend that a single, monolithic evaluation method (\textit{e.g.}, a “Pure LMM” score) is insufficient to capture the full spectrum of image editing quality. The 30 dimensions in our benchmark are decoupled, targeting distinct facets of quality ranging from objective technical fidelity to complex cognitive reasoning.

To achieve the highest possible alignment with human judgment, we designed a hybrid evaluation system that explicitly matches the evaluation needs of each dimension to the most appropriate tool. Our methodology is built on a “best-tool-for-the-job” principle, which can be categorized into three distinct evaluation pathways.

\subsection{Specialist Models: For Objective and Perceptual Quantification}
\label{ssec:specialist_models}

For well-defined, global image properties, mature specialist models provide the most stable, objective, and unbiased scores.

\paragraph{Dimensions:} \texttt{Overall Image Quality}, \texttt{Aesthetic Quality}.

\paragraph{Method:} Specialist Models (Q-Insight, ArtiMuse).

\paragraph{Rationale:} These dimensions require assessing global, technical (\textit{e.g.}, clarity, noise) or artistic (\textit{e.g.}, composition, color harmony) quality. Unlike an LMM, whose judgment (as in our “Pure LMM” baseline) can be easily biased by the *semantic content* of an edit (\textit{e.g.}, an LMM might give a high score to a “semantically correct” but blurry edit), these specialist tools are trained on specific, large-scale datasets (like Koniq-10k) to provide consistent quantitative scores for these precise attributes, aligning closely with human perception of these specific factors.

\subsection{Hybrid LMM-Specialist Pipelines: For High-Fidelity and Content-Specific Tasks}
\label{ssec:hybrid_pipelines}

This hybrid approach is a core innovation of I2I-Bench. It combines the semantic understanding of LMMs (“what to look at”) with the precise quantification of specialist tools (“how to measure it”), overcoming the limitations of using either one alone.

\subsubsection{Hybrid OCR: For Textual Accuracy}
\label{sssec:hybrid_ocr}

\paragraph{Dimensions:} \texttt{Text Content \& Style Accuracy}.

\paragraph{Method:} Hybrid LMM VQA-5Level + Specialist OCR (GOT-OCR2.0).

\paragraph{Rationale:} We do not use a “Pure LMM” (baseline) for this dimension due to its poor performance in precise OCR. LMMs frequently “hallucinate”---misreading, omitting, or inventing text. This dimension involves two distinct sub-tasks: (1) \textbf{Content:} Is the spelling correct? (2) \textbf{Style:} Are the font, position, and color correct?
\begin{itemize}
    \item For \textbf{Content (1)}, a binary, objective task, the specialist OCR model (GOT-OCR2.0) provides a “ground truth” score for content accuracy ($S_c$).
    \item For \textbf{Style (2)}, a subjective, semantic judgment, the LMM is the ideal tool (via LMM VQA-5Level, yielding $S_s$).
\end{itemize}
Our hybrid pipeline (Eq. 5) uses the objective OCR score to “gate” the subjective LMM style score, ensuring a model does not receive a high score for generating beautifully styled but \textit{incorrectly spelled} text.

\paragraph{Further Justification for the Fusion Strategy:}
We select the piecewise function in Eq.4 (from the main paper) based on a careful consideration of human perceptual mechanisms, rather than a simpler combination.

\begin{itemize}
    \item \textbf{Inadequacy of Additive Fusion:} A simple additive combination (\textit{e.g}., $S_c + S_s$) is fundamentally unsuitable. It fails to implement a “gating” mechanism, meaning a perfect style score ($S_s=1.0$) could erroneously compensate for completely incorrect content (\textit{e.g.}, $S_c=0.1$), leading to a high score for a failed edit.

    \item \textbf{Why Piecewise is Superior to Simple Multiplication:} As you correctly noted, a simple multiplicative fusion (\textit{e.g.}, $S_c \times S_s$) does provide a basic gating effect. However, we found it insufficient as it fails to capture the \textit{non-linear} nature of human perception. Our empirical observations suggest that human evaluators do not assess textual accuracy on a continuous linear scale. Instead, they tend to “bucket” the results into coarse-grained categories:
    \begin{enumerate}
        \item Perfectly Correct ($S_c \approx 1.0$)
        \item Mostly Correct / Minor Error (\textit{e.g.}, $S_c \ge 0.8$)
        \item Partially Correct / Significant Errors (\textit{e.g.}, $S_c \approx 0.5$)
        \item Completely Wrong ($S_c < 0.3$)
    \end{enumerate}
    A simple multiplication treats the difference between $S_c=0.9$ and $S_c=0.8$ the same as the difference between $S_c=0.6$ and $S_c=0.5$. In contrast, our piecewise function is explicitly designed to model these discrete human perceptual thresholds, applying a gentle penalty for “mostly correct” results while applying a severe penalty once the accuracy drops below a “failure” threshold.

    \item \textbf{Disparity in Task Objectivity:} This design also accounts for the different nature of the sub-tasks. Rendering correct content ($S_c$) is an objective, difficult, and high-stakes task strictly measured by OCR. Rendering style ($S_s$) is a more subjective and, comparatively, lower-difficulty task evaluated by the LMM. The objective $S_c$ score must therefore serve as a robust, non-negotiable filter for the subjective $S_s$ score.

    \item \textbf{Empirical Validation:} The specific thresholds (\textit{e.g.}, 0.8, 0.6) and their corresponding multipliers (0.8, 0.5, 0.1) were chosen empirically. During our development, we tested several fusion configurations (including simple multiplication) and found that this specific piecewise setup yielded the \textbf{highest Pearson correlation} with our human preference annotations, validating its superior alignment with human judgment.
\end{itemize}

\subsubsection{Hybrid Feature Matching: For Fidelity and Identity Preservation}
\label{sssec:hybrid_feature}

\paragraph{Dimensions:} \texttt{Non-Edited Region Fidelity}, \texttt{Subject Identity Fidelity}, \texttt{Subject Consistency}, \texttt{Detail Fidelity/Preservation}.

\paragraph{Method:} Hybrid LMM-Specialist (Sa2VA + DINOv3).

\paragraph{Rationale:} We do not use a “Pure LMM” (baseline) for these dimensions due to “semantic drift.” An LMM judges high-level \textit{concepts} (\textit{e.g.}, it might consider a \textit{different person} in the same clothes as having high “subject identity”), not true \textit{perceptual similarity}.
\begin{itemize}
    \item Conversely, simple pixel-level metrics (\textit{e.g.}, PSNR or LPIPS) fail to capture \textit{feature-level} identity.
    \item Our hybrid pipeline (Eq. 2) leverages the LMM (via Sa2VA) for its strength: \textbf{semantic segmentation} (to identify \textit{which} pixels correspond to the “subject” or “non-edited background”).
    \item It then leverages the specialist model (DINOv3) for its strength: \textbf{feature-level comparison} (extracting and comparing feature vectors from those segmented regions). This provides a quantitative, robust score for fidelity that neither an LMM nor a simple metric could achieve.
\end{itemize}

\subsection{Generalist LMM Pipelines: For Semantic and Cognitive Judgments}
\label{ssec:generalist_lmm}

For dimensions where the core task is semantic understanding, logical reasoning, or subjective assessment, the LMM is the ideal, and often only, tool. We further refine this by selecting different LMM VQA structures based on the \textit{nature} of the judgment required.

\subsubsection{5-Level VQA (for Subjective, Holistic Scoring)}
\label{sssec:5level_vqa}

\paragraph{Dimensions:} \texttt{Blending Naturalness}, \texttt{Editing Artifacts}, \texttt{Instruction Following (Macro)}.

\paragraph{Method:} LMM VQA-5Level (Eq. 1).

\paragraph{Rationale:} These dimensions are inherently \textbf{subjective and holistic}. There is no “binary” correct answer for \texttt{Blending Naturalness} or \texttt{Editing Artifacts}; they exist on a spectrum. The 5-level weighted score (from “excellent” to “bad”) is designed to capture this nuanced, perceptual “feel.” Similarly, \texttt{Instruction Following (Macro)} assesses the overall \textit{gist} and \textit{intent} of the edit, making it a perfect choice for a 5-level holistic judgment.

\subsubsection{2-Level VQA (for Binary, Factual Success)}
\label{sssec:2level_vqa}

\paragraph{Dimensions:} \texttt{Object Manipulation Accuracy}, \texttt{Local Attribute Accuracy}, \texttt{Action/State Change Accuracy}, \texttt{Spatial Accuracy}, \texttt{World Knowledge \& Reasoning (SE)}.

\paragraph{Method:} LMM VQA-2Level (Eq. 4).

\paragraph{Rationale:} This targets factual, binary (Yes/No) task success, primarily in Single-Image (SE) edits. The edit either \textit{happened as specified} (Yes) or \textit{it did not} (No).
\begin{itemize}
    \item “Was the horse changed to \textit{white}?” (Yes/No).
    \item “Was the \textit{correct} mayor's statue added?” (Yes/No).
    \item “Was the object placed to the \textit{left} of the target?” (Yes/No).
\end{itemize}
A 2-Level (Yes/No) VQA is the most direct and unambiguous way to measure this \textbf{knowledge-retrieval} or \textbf{instruction-execution} semantic correctness. \textbf{Crucially, we are not testing the LMM's own knowledge; we use the LMM VQA to judge if the “edited image” matches our “pre-defined correct answer set.” This ensures the objectivity of the evaluation.}

\subsubsection{Multi-Question VQA (for Complex, Decomposed Reasoning)}
\label{sssec:multi_question_vqa}

\paragraph{Dimensions:} \texttt{Physical Plausibility}, \texttt{Composition \& Interaction}.

\paragraph{Method:} LMM Multi-Question VQA (Eq. 3).

\paragraph{Rationale:} These are cognitively complex dimensions. Asking an LMM for a single 1-5 score (i.e., the “Pure LMM” baseline) is \textbf{unreliable}. An LMM suffers from “attention bias” when evaluating complex scenes; it may focus only on the edited subject while completely ignoring its incorrect shadow or perspective.
\begin{itemize}
    \item Our Multi-Question VQA method acts as a \textbf{“forced attention mechanism.”} It \textit{decomposes} the complex concept into a series of simple, verifiable sub-questions (\textit{e.g.}, “1. Is the shadow direction of the new object correct?” “2. Is the perspective of the new object consistent?”).
    \item This forces the LMM to perform a more robust, “chain-of-thought”-like analysis across multiple facets (lighting, shadows, perspective, occlusion). The resulting score, an aggregation of “Yes” answers (Eq. 3), is far more reliable and fine-grained than a single, potentially biased, holistic judgment.
\end{itemize}

\subsubsection{LMM VQA (for Multi-Image Abstract Cognition \& Cross-Image Relations)}
\label{sssec:multi_image_vqa}

These are the most difficult, SOTA-challenging dimensions in I2I-Bench, characterized by their need for \textbf{relational understanding} and \textbf{semantic reasoning across multiple source images}. No specialist model can perform such tasks, making the LMM the only tool for evaluation. We match the VQA structure to the nature of each task:

\paragraph{Dimension 1: \texttt{Cross-Source Attribute/Pose Transfer}}
\begin{itemize}
    \item \textbf{Method:} LMM VQA-2Level (Binary Factual Judgment).
    \item \textbf{Rationale:} This task (\textit{e.g.}, “Transfer pose from A in Image 1 to B in Image 2”) is a purely \textbf{“relational instruction.”} The LMM must: (1) identify the source attribute (pose) in Image 1; (2) identify the target (B) in Image 2; and (3) judge if B in the generated image has \textit{factually adopted} A's pose. No specialist model (\textit{e.g.}, pose-estimator) can understand the semantic command “to transfer.” This is a binary (Yes/No) cross-image semantic verification, making 2-Level VQA most appropriate.
\end{itemize}

\paragraph{Dimension 2: \texttt{Inter-Subject Consistency}}
\begin{itemize}
    \item \textbf{Method:} LMM VQA-5Level (Subjective Spectral Judgment).
    \item \textbf{Rationale:} This task assesses “how harmonious subjects from different sources look when composited together.” This is distinct from Sec 2.2's \texttt{Subject Consistency} (which measures fidelity to the \textit{source}). This is a high-level, \textbf{scene-wide artistic and semantic judgment}. The LMM must evaluate if lighting, scale, and artistic style (\textit{e.g.}, a photorealistic person vs. an anime person) are consistent \textit{within the new scene}. This is a subjective “feel”, not a binary (Yes/No) question, making the 5-Level VQA spectrum the best fit.
\end{itemize}

\paragraph{Dimension 3: \texttt{World Knowledge \& Reasoning (ME)}}
\begin{itemize}
    \item \textbf{Method:} LMM VQA-2Level (Binary Factual Judgment).
    \item \textbf{Rationale:} This task, unlike its SE “knowledge retrieval” counterpart, demands \textbf{“abstract logical reasoning”} (\textit{e.g.}, map inference, Sudoku solving, logical combinations based on multiple images). The task occurs in \textit{logical space}, not pixel space. The LMM is the only tool that can understand “logic”. In line with Sec 3.2, to ensure objectivity, we \textbf{use the LMM VQA to judge if the “edited image” satisfies our “pre-defined correct answer set”} (\textit{e.g.}, the correct Sudoku solution, the correct city name from the map). This makes the evaluation an objective (Yes/No) check, for which 2-Level VQA is ideal.
\end{itemize}

\section{Prompt Quota per I2I-Bench Evaluation Dimension}
\label{sec:prompt quota}

\noindent This section details the exact number of prompts used to calculate the final scores for each model across all 30 dimensions in the proposed I2I-Bench evaluation suite. The total number of prompts is 1000, split equally between 500 Single-Image Editing (SE) prompts and 500 Multi-Image Editing (ME) prompts.

\label{sec:se-dims}

\begin{table}[h!]
\centering
\caption{Prompt Quota for Single-Image Editing (SE) Dimensions.}
\label{tab:se-quota}
\begin{tabular}{|l|c|}
\hline
\textbf{English Dimension Name} & \textbf{Prompt Quota} \\
\hline
Aesthetic Quality & 500 \\
Blending Naturalness & 500 \\
Editing Artifacts & 500 \\
Image Quality & 500 \\
Instruction Following (Macro) & 500 \\
Non-Edited Element Fidelity & 500 \\
Physical Plausibility & 500 \\
Composition \& Interaction & 350 \\
Object Manipulation Accuracy & 332 \\
Local Attribute Accuracy & 282 \\
Spatial Accuracy & 275 \\
Subject Identity Fidelity & 233 \\
Text Content \& Style Accuracy & 100 \\
World Knowledge \& Reasoning & 100 \\
Action/State Change Accuracy & 86 \\
\hline
\end{tabular}
\end{table}

\label{sec:me-dims}

\begin{table}[h!]
\centering
\caption{Prompt Quota for Multi-Image Editing (ME) Dimensions.}
\label{tab:me-quota}
\begin{tabular}{|l|c|}
\hline
\textbf{English Dimension Name} & \textbf{Prompt Quota} \\
\hline
Aesthetic Quality & 500 \\
Blending Naturalness & 500 \\
Composition \& Interaction & 500 \\
Detail Fidelity/Preservation & 500 \\
Image Quality & 500 \\
Instruction Following (Macro) & 500 \\
Non-Edited Element Fidelity & 500 \\
Physical Plausibility & 500 \\
Subject Consistency & 500 \\
Subject Extraction \& Composition & 500 \\
Spatial Accuracy & 476 \\
Inter-Subject Consistency & 412 \\
Text Content \& Style Accuracy & 200 \\
Cross-Source Attribute/Pose Transfer & 126 \\
World Knowledge \& Reasoning & 100 \\
\hline
\end{tabular}

\end{table}

\section{Comparative Analysis with LMM4Edit}

To demonstrate the superiority of our proposed evaluation suite, we conducted a comparative analysis against LMM4Edit, a recent image editing evaluation metric based on LMMs. We performed inference using LMM4Edit on the Single-Image Editing component of I2I-Bench. While LMM4Edit provides pre-trained weights corresponding to multiple dimensions, we observed that not all checkpoints were fully applicable within our testing environment due to technical inconsistencies. Consequently, we selected one of the viable weight versions to conduct the comparative experiment.

Table~\ref{tab:lmm4edit_comparison_en} presents a detailed comparison of Pearson's Rho correlations between LMM4Edit and our method (Ours). The results unequivocally demonstrate that I2I-Bench significantly outperforms LMM4Edit across the vast majority of evaluation dimensions. Specifically:

\begin{itemize}
    \item \textbf{Superior Overall Alignment}: Our method achieves a remarkably high average correlation of \textbf{0.9425} (excluding nan), compared to 0.5968 for LMM4Edit. This substantial gap validates the effectiveness of our "Specialist-Generalist Hybrid" evaluation strategy in aligning with human perception.
    \item \textbf{Robustness in Fundamental Dimensions}: LMM4Edit exhibits critical failures in fundamental quality assessment. Notably, it shows a negative correlation (-0.4568) in \textit{Blending-Naturalness} and a weak correlation (0.2508) in \textit{Image-Quality}, failing to correctly penalize artifacts. In contrast, our method achieves high consistency scores of 0.8866 and 0.9033, respectively, in these dimensions.
    \item \textbf{Reasoning Capabilities}: In complex tasks such as \textit{World-Knowledge-Reasoning} and \textit{Text-Content-Style-Accuracy}, our method demonstrates overwhelming superiority due to the integration of specialized tools (OCR and VQA specialists), whereas the pure LMM-based approach of LMM4Edit struggles significantly.
\end{itemize}

In conclusion, this experiment confirms that I2I-Bench provides a far more robust, accurate, and human-aligned evaluation framework compared to existing LMM-based metrics.

\begin{table}[htbp]
    \centering
    \caption{Comparison of Pearson's Rho correlation with human preference between LMM4Edit and our method (Ours) on the I2I-Bench Single-Image Editing task. Our method demonstrates significant superiority across almost all dimensions.}
    \label{tab:lmm4edit_comparison_en}
    \resizebox{\linewidth}{!}{%
    \begin{tabular}{lccc}
        \toprule
        \textbf{Dimension} & \textbf{LMM4Edit ($\rho$)} & \textbf{Ours ($\rho$)} & \textbf{Gap ($\Delta$)} \\
        \midrule
        Image-Quality & 0.2508 & \textbf{0.9033} & +0.6525 \\
        Aesthetic-Quality & 0.7846 & \textbf{0.9889} & +0.2043 \\
        Blending-Naturalness & -0.4568 & \textbf{0.8866} & +1.3434 \\
        Non-Edited-Element-Fidelity & 0.5984 & \textbf{0.9019} & +0.3035 \\
        Subject-Identity-Fidelity & 0.8210 & \textbf{0.9133} & +0.0923 \\
        Physical-Plausibility & 0.7683 & \textbf{0.8055} & +0.0372 \\
        Editing-Artifacts & 0.6551 & \textbf{0.9006} & +0.2455 \\
        Instruction-Following-Macro & 0.6266 & \textbf{0.9960} & +0.3694 \\
        Object-Manipulation-Accuracy & 0.7559 & \textbf{0.9787} & +0.2228 \\
        Local-Attribute-Accuracy & 0.7445 & \textbf{0.9877} & +0.2432 \\
        Action-State-Change-Accuracy & 0.5996 & \textbf{0.9839} & +0.3843 \\
        Spatial-Accuracy & \textbf{0.9655} & 0.9303 & -0.0352 \\
        Text-Content-Style-Accuracy & 0.5873 & \textbf{0.9979} & +0.4106 \\
        World-Knowledge-Reasoning & 0.2885 & \textbf{0.9628} & +0.6743 \\
        Composition-Interaction & 0.9631 & \textbf{0.9997} & +0.0366 \\
        \midrule
        \textbf{Average Correlation} & 0.5968 & \textbf{0.9425} & \textbf{+0.3457} \\
        \bottomrule
    \end{tabular}%
    }
\end{table}

\section{Evaluation Pipeline and Prompt Details}

This section details the automated hybrid evaluation methods used to assess the 30 fine-grained dimensions in I2I-Bench. The “Generalist” Large Multimodal Model (LMM) used for all evaluations is Qwen3-VL-8-Instruct.

\subsection{Single-Reference (SE) Evaluation Dimensions}

\subsubsection{SE: Common Dimensions}
These 7 dimensions assess the fundamental quality of all Single-Reference (SE) editing tasks.

\paragraph{1. Overall Image Quality \& 2. Aesthetic Quality}
\textbf{Pipeline Type:} Specialist Models.
\textbf{Tools:} Q-Insight, ArtiMuse.
\textbf{Prompts:} N/A (Scores are obtained directly from the Specialist models).

\paragraph{3. Blending Naturalness}
\textbf{Pipeline Type:} LMM VQA-5Level.
\textbf{Tools:} Qwen3-VL-8-Instruct.

\noindent\textbf{System Prompt:}

\begin{promptbox}
You are an expert evaluator of image photorealism and coherence. Your specific task is to assess the realism of the edit itself. Evaluate how seamlessly the modified or added elements integrate with the rest of the image in terms of lighting, shadows, perspective, and texture. A high rating means the final image looks natural and plausible, as if it were a single, untouched photograph. Do not focus on whether the instruction was followed literally. Your response must be one of the following five words directly: excellent, good, fair, poor, bad. Do not add any introductory phrases.
\end{promptbox}

\noindent\textbf{User Prompt ($Q$):}

\begin{promptbox}
Please evaluate the realism and coherence of the edit in the 'Generated Image' compared to the 'Source Images'. Assess how seamlessly the edited elements integrate with the rest of the image in terms of lighting, shadows, and overall plausibility. Your response must begin with one of the five rating words: excellent, good, fair, poor, bad.
\end{promptbox}

\paragraph{4. Generative/Editing Artifacts}
\textbf{Pipeline Type:} LMM VQA-5Level.
\textbf{Tools:} Qwen3-VL-8-Instruct.

\noindent\textbf{System Prompt:}

\begin{promptbox}
You are an expert evaluator of image editing quality. You will be given an 'Original Image', an 'Editing Instruction', and the resulting 'Edited Image'. Your task is to assess the 'Edited Image' for *unwanted artifacts* introduced *during the editing process*, not artifacts that were in the original.

**Crucially**: If the instruction is stylistic (\textit{e.g.}, 'turn into a painting', 'make it look like Van Gogh'), *do not* penalize the image for looking 'unnatural'. Instead, judge if the *application* of the style is flawed (\textit{e.g.}, distorted, incomplete, blotchy).

Also, heavily penalize edits that *fail to preserve* unedited regions (\textit{e.g.}, if the instruction is 'change the woman's hat', her face and the background should remain unchanged).

Based on the *severity* of these *unwanted* artifacts, rate the 'Edited Image' using *only* one of the following five words:
1. excellent (Flawless edit. The instruction is followed perfectly with *zero* unwanted artifacts. Unedited areas are perfectly preserved.)
2. good (Minor, hard-to-notice artifacts. The edit is successful but may have tiny imperfections upon close inspection.)
3. fair (Noticeable artifacts. The edit is mostly successful, but there are visible flaws like slight warping, minor texture loss, or imperfect blending.)
4. poor (Significant, obvious artifacts. The edit is flawed, with clear distortions, unnatural warping, or significant damage to unedited areas.)
5. bad (Extreme, severe artifacts. The edit is a total failure, resulting in a grotesque, distorted, or nonsensical image.)
\end{promptbox}

\noindent\textbf{User Prompt ($Q$):}

\begin{promptbox}
Based on the instruction, how severe are the *unwanted artifacts* in the 'Edited Image'?
\end{promptbox}

\paragraph{5. Instruction Following (Macro)}
\textbf{Pipeline Type:} LMM VQA-5Level.
\textbf{Tools:} Qwen3-VL-8-Instruct.

\noindent\textbf{System Prompt:}

\begin{promptbox}
You are a meticulous evaluator specializing in text-to-image editing. Your sole task is to assess how accurately the edited image reflects the given instruction, based on the original image. Focus exclusively on whether the edit described in the instruction was performed correctly. You must ignore all other factors, such as overall image quality or any unintended changes in areas not mentioned in the instruction. Your response must be one of the following five words directly: excellent, good, fair, poor, bad. Do not add any introductory phrases.
\end{promptbox}

\noindent\textbf{User Prompt ($Q$):}

\begin{promptbox}
Please evaluate whether the 'Generated Image' successfully implements the following instruction. Instruction: “{prompt}”. Your response must begin with one of the five rating words: excellent, good, fair, poor, bad.
\end{promptbox}

\paragraph{6. Non-Edited Region Fidelity}
\textbf{Pipeline Type:} Hybrid LMM-Specialist.
\textbf{Tools:} LMM (Qwen3-VL-8-Instruct) + Sa2VA + DINOv3.
\textbf{Description:} The LMM generates a segmentation command for the \textbf{''edited''} region. The mask is then \textbf{inverted (NOT)} to isolate the \textit{non-edited region}.

\noindent\textbf{LMM Segmentation Command Prompt ($Q$):}

\begin{promptbox}
You are an image segmentation assistant. Compare <image\_1> (Original Image) and <image\_2> (Edited Image), and considering the editing instruction: “{edit\_instruction}”, generate a text command to segment the edited or modified region in the image. Your response MUST begin with “Please segment”. This command will be used to segment the edited image in isolation. Therefore, ensure the command is clear, focuses only on the edited image, and makes no reference to the original image, as this would confuse the segmentation model.
\end{promptbox}

\paragraph{7. Physical Plausibility}
\textbf{Pipeline Type:} LMM Multi-Question VQA (Multi-VQA).
\textbf{Tools:} Qwen3-VL-8-Instruct.

\noindent\textbf{QG (Question Generation) Prompt:}

\begin{promptbox}
You are an expert in physical plausibility evaluation. Your task is to design a set of (5 to 7) precise, specific binary (Yes/No) questions based on the given original image and an editing instruction, to evaluate whether the edited image is physically plausible.

When designing questions, strictly follow this thought process and principles:

1.  **Analyze Scene \& Instruction**:
    * **Identify Subject \& Action**: What object (subject) is being manipulated (added, removed, replaced, attribute changed)?
    * **Analyze Physical Context**: Observe the original image to identify relevant physical properties and environmental factors: \textit{e.g.}, lighting direction, shadows, reflective surfaces, support relationships, rigid vs. soft bodies.

2.  **Generate Questions**: Based on the analysis, generate questions from the following physical dimensions. Each must be a closed-ended Yes/No question.
    * **Optics (Shadows, Reflections)**: If an object was added/moved, is its new shadow consistent with the scene's light source? If an object was removed, is its shadow also gone? Are reflections on nearby surfaces correctly updated?
    * **Mechanics (Deformation, Support)**: If a heavy object is placed on a soft surface, does the surface show plausible deformation? If a supporting object is removed, does the object it supported (\textit{e.g.}, a vase on a table) defy gravity?
    * **State Transition (Weather, State)**: If the instruction changes the weather (\textit{e.g.}, 'make it winter'), are all elements (trees, ground) consistently updated (\textit{e.g.}, covered in snow)?

3.  **Output Format**:
    Please output a JSON-formatted string containing a list of questions. The JSON object should have a key named “physical\_questions” with a value that is a list of strings.
    Example:
    \{
      “physical\_questions”: [
        “Question 1: ...?”,
        “Question 2: ...?”,
        “Question 3: ...?”
      ]
    \}

Now, please generate physical plausibility evaluation questions based on the following image and instruction.
Instruction: “{prompt}”.
\end{promptbox}

\noindent\textbf{QA (Question Answering) Prompt:}

\begin{promptbox}
You are an expert in physical plausibility evaluation. Please carefully observe the provided image and judge based on the following question.

Question: “{question}”

Please answer with only one word based on visual evidence: Yes or No.
\end{promptbox}

\subsubsection{SE: Specific Dimensions}
These 8 dimensions assess the execution quality of specific instructions.

\paragraph{8--11. Object Manipulation, Local Attribute, Action/State Change, Spatial Accuracy}
\textbf{Pipeline Type:} LMM VQA-2Level.
\textbf{Tools:} Qwen3-VL-8-Instruct.

\noindent\textbf{System Prompt:}

\begin{promptbox}
You are an expert in evaluating image editing. Your task is to determine if the edited image successfully implements the given instruction compared to the original image. Focus solely on whether the instruction was achieved. Answer strictly and only with 'Yes' or 'No'.
\end{promptbox}

\noindent\textbf{User Prompt ($Q$):}

\begin{promptbox}
Instruction: “{prompt}”. Does Image 2 successfully implement the instruction compared to Image 1? Answer Yes or No.
\end{promptbox}

\paragraph{12. Text Content \& Style Accuracy}
\textbf{Pipeline Type:} Multi-step Hybrid Pipeline.
\textbf{Tools:} GOT-OCR2.0 + LMM (Qwen3-VL-8-Instruct).
\textbf{Description:} GOT-OCR2.0 evaluates content accuracy ($S_c$). The LMM then uses a VQA-5Level pipeline to evaluate style and position ($S_s$).

\noindent\textbf{LMM Style/Position Prompt ($Q$):}

\begin{promptbox}
You are an expert evaluator for AI-generated images, specializing in text rendering. Your task is to evaluate how well an image follows the text-related **style and position** instructions from a user prompt.

**CRITICAL RULE:** Do NOT evaluate the text's spelling or accuracy. Assume the spelling is correct, even if it is not. Your score must ONLY reflect the non-accuracy requirements (like font style, color, placement, etc.).

You will be given:
1.  **[User Prompt]**: The original prompt used to generate the image.
2.  **[Target Text]**: The specific text string that was requested.
3.  **[Image]**: The generated image.

**Step 1: Analyze Requirements**
First, analyze the [User Prompt] to identify the specific requirements for:
* **Text Style**: What instructions were given for the text's appearance (\textit{e.g.}, “neon”, “handwritten”, “bold”, “red color”, “glowing”, “artistic font”)?
* **Text Position**: What instructions were given for the text's location (\textit{e.g.}, “on the sign”, “in the top-left corner”, “on the t-shirt”)?

**Step 2: Evaluate Image against Requirements**
Compare the text in the [Image] against the non-accuracy requirements you identified. Remember to IGNORE spelling errors.

**Step 3: Assign a Single Score (1-5)**
Provide a single, holistic score for **Style and Position Compliance** based on this rubric.
* [5] Excellent Match: All specified style and position instructions were followed perfectly.
* [4] Good Match: All specified instructions were followed, but with minor deviations.
* [3] Partial Match: The core idea of *at least one* instruction was attempted but executed poorly, OR one major instruction was followed while another was missed.
* [2] Poor Match: At least one specified instruction was clearly ignored or failed.
* [1] No Match: All specified style and position instructions were completely ignored.

**Step 4: Provide Output in JSON Format**
Provide your evaluation in a strict JSON format. Do not include any text outside the JSON block.

**JSON Output Format:**
\{
  “analysis”: \{
    “style\_requirement”: “...”,
    “position\_requirement”: “...”,
    “image\_observation”: “...”,
    “reasoning”: “...”,
    “score”: [1-5]
  \}
\}
\end{promptbox}

\paragraph{13. World Knowledge \& Reasoning}
\textbf{Pipeline Type:} LMM VQA-2Level.
\textbf{Tools:} Qwen3-VL-8-Instruct.

\noindent\textbf{System Prompt:}

\begin{promptbox}
You are an expert evaluator for image editing based on multiple reference images. Your task is to determine if the 'Edited Image' successfully implements the requested change described in the 'Instruction', based *specifically* on the 'Evaluation Criteria (Hint)'. Compare the 'Reference Images' and 'Edited Image'. Your response must be *only* the single word 'Yes' or 'No'.

'Yes' = The edit was successfully implemented according to the hint.
'No' = The edit was not successfully implemented according to the hint.
\end{promptbox}

\noindent\textbf{User Prompt ($Q$):}

\begin{promptbox}
**Instruction (Prompt):** {prompt}

**Evaluation Criteria (Hint):** {hint}

Based on ALL reference images, the instruction, and the specific criteria in the hint, has the edit been successfully implemented in the 'Edited Image'? Answer with only 'Yes' or 'No'.
\end{promptbox}

\paragraph{14. Subject Identity Fidelity}
\textbf{Pipeline Type:} Hybrid LMM-Specialist.
\textbf{Tools:} LMM (Qwen3-VL-8-Instruct) + Sa2VA + DINOv3.
\textbf{Description:} The LMM generates a command to segment regions that \textbf{''should remain unchanged''} (\textit{e.g.}, the face, if the instruction is 'change the shirt'). The mask is \textbf{not inverted}.

\noindent\textbf{LMM Segmentation Command Prompt ($Q$):}

\begin{promptbox}
You are an expert in image editing analysis. Given an original image, an edited image, and the editing instruction, identify all distinct main subjects or regions present in the original image that *should remain unchanged* according to the instruction.

Your output MUST BE ONLY a JSON list of strings. Each string must be a separate segmentation command for one distinct subject/region, in the format 'Please segment [subject name in English]'.

For example: [“Please segment background sky”, “Please segment mountains”, “Please segment main building”]
\end{promptbox}

\paragraph{15. Composition \& Interaction}
\textbf{Pipeline Type:} LMM Multi-Question VQA.
\textbf{Tools:} Qwen3-VL-8-Instruct.

\noindent\textbf{QG (Question Generation) Prompt:}

\begin{promptbox}
You are an expert in visual arts and composition evaluation. Your task is to design a set of (5 to 7) precise, specific binary (Yes/No) questions based on the given original image and an editing instruction, to evaluate whether the edited image is plausible in terms of **composition, perspective, scale, and interaction**.

When designing questions, strictly follow this thought process and principles:

1.  **Analyze Scene \& Instruction**:
    * **Identify Subject \& Action**: What object (subject) is being manipulated (added, removed, replaced, attribute/pose changed)?
    * **Analyze Visual Context**: Observe the original image to identify key visual elements: scene perspective (close-up, long-shot, eye-level), key objects, spatial layout (foreground/background), and interaction area implied by the instruction.

2.  **Generate Questions**: Based on the analysis, generate questions from the following dimensions.
    * **Composition \& Placement**: Is the new/moved object in a logical position? Is its occlusion (in front of/behind other objects) correct?
    * **Perspective \& Scale**: Is the scale of the new/modified object consistent with other objects in the scene? Does its perspective match the scene's perspective?
    * **Interaction \& Naturalness**: If a pose was changed, is it anatomically natural? If objects are interacting (\textit{e.g.}, hand holding a balloon), is the contact point believable?

3.  **Output Format**:
    Please output a JSON-formatted string containing a list of questions. The JSON object should have a key named “composition\_questions” with a value that is a list of strings.

Now, please generate composition and interaction plausibility evaluation questions based on the following image and instruction.
Instruction: “{prompt}”.
\end{promptbox}

\noindent\textbf{QA (Question Answering) Prompt:} (Reused from SE Physical Plausibility).

\subsection{Multi-Reference (ME) Evaluation Dimensions}
These 15 dimensions evaluate complex multi-image editing tasks.

\subsubsection{ME: Common Dimensions}
These 9 dimensions are shared with the SE category.

\paragraph{1--5. Aesthetic Quality, Blending Naturalness, Editing Artifacts, Image Quality, Instr. Following (Macro)}
\textbf{Prompts:} Reused from the corresponding SE definitions.

\paragraph{6. Spatial Accuracy}
\textbf{Prompts:} Reused from the SE LMM VQA-2Level definition.

\paragraph{7. Composition \& Interaction}
\textbf{Pipeline Type:} LMM Multi-Question VQA.

\noindent\textbf{QG Prompt (ME-Specific):}

\begin{promptbox}
You are an expert in visual arts and composition evaluation. Your task is to design a set of (5 to 7) precise, specific binary (Yes/No) questions based on *multiple* given source images (labeled Figure 1, Figure 2, ...) and an editing instruction, to evaluate whether the *final composited image* is plausible in terms of **composition, perspective, scale, and interaction**.

When designing questions, strictly follow this thought process and principles:

1.  **Analyze Scene \& Instruction**:
    * **Identify Sources \& Composition**: What elements are extracted from which images? How are they combined, modified, and placed?
    * **Analyze Visual Context**: Check perspective, scale, and interaction. Does the instruction require interaction (\textit{e.g.}, A sits on B)?

2.  **Generate Questions**:
    * **Composition \& Placement**: Are elements placed logically (\textit{e.g.}, not floating)? Is occlusion correct?
    * **Perspective \& Scale**: Is the scale of an element from Fig 1 consistent with the scene from Fig 2? Do all elements share a consistent perspective?
    * **Interaction \& Naturalness**: If a pose was changed to interact (\textit{e.g.}, sit on a chair), is the final pose natural? Is the physical contact believable?

3.  **Output Format**:
    Please output a JSON-formatted string... The JSON object should have a key named “composition\_questions”...

Now, please generate... questions based on the following *multiple* images and the instruction.
Instruction: “{prompt}”.
\end{promptbox}

\noindent\textbf{QA Prompt:} (Reused from SE Physical Plausibility).

\paragraph{8. Physical Plausibility}
\textbf{Pipeline Type:} LMM Multi-Question VQA.

\noindent\textbf{QG Prompt (ME-Specific):}

\begin{promptbox}
You are an expert in physical plausibility evaluation. Your task is to design a set of (5 to 7) precise, specific binary (Yes/No) questions based on *multiple* given source images (labeled Figure 1, Figure 2, ...) and an editing instruction, to evaluate whether the *final composited image* is physically plausible.

When designing questions, strictly follow this thought process and principles:

1.  **Analyze Scene \& Instruction**:
    * **Identify Sources \& Composition**: What elements are extracted and how are they combined?
    * **Analyze Physical Context**: Are the lighting, shadows, and physics consistent *between* elements from different sources?
    * **Analyze Interaction**: Do elements interact? Do these interactions obey physical laws (support, occlusion, deformation)?

2.  **Generate Questions**:
    * **Optics (Lighting/Shadows)**: Are the shadows and lighting on all combined elements consistent with a single, unified light source?
    * **Mechanics (Support/Deformation)**: If an element from Fig 1 is placed on a soft element from Fig 2, does the surface plausibly deform? Are support structures logical?
    * **State Consistency**: If the instruction changes the global state (\textit{e.g.}, 'make it rain'), does this state apply consistently to all elements from all sources?

3.  **Output Format**:
    Please output a JSON-formatted string... The JSON object should have a key named “physical\_questions”...

Now, please generate... questions based on the following *multiple* images and the instruction.
Instruction: “{prompt}”.
\end{promptbox}

\noindent\textbf{QA Prompt:} (Reused from SE Physical Plausibility).

\paragraph{9. Non-Edited Element Fidelity}
\textbf{Pipeline Type:} Hybrid LMM-Specialist.

\noindent\textbf{LMM Segmentation Command Prompt ($Q$):}

\begin{promptbox}
You are an expert in analyzing complex image editing instructions. Given multiple source images, an edited image, and the instruction, your task is to identify which specific *source subjects* (\textit{e.g.}, 'Luffy from Figure 2', 'Conan from Figure 3') are explicitly instructed to be preserved *without changes* (\textit{e.g.}, 'keep their original poses', 'remain unchanged').

For each such non-edited subject you find, output a JSON object containing:
1. 'source\_index': The 0-based index of the source image where this subject originates.
2. 'segmentation\_prompt': A short segmentation command in the format 'Please segment [subject name in English]'.

Your output MUST BE ONLY a JSON list of these objects.
Example: [{“source\_index”: 1, “segmentation\_prompt”: “Please segment Luffy”}, {“source\_index”: 2, “segmentation\_prompt”: “Please segment Conan”}]
If the instruction modifies *all* subjects in some way (\textit{e.g.}, 'put all in new clothes'), output an empty list `[]`.
Do not add any text before or after the JSON list.
\end{promptbox}

\subsubsection{ME: Specific Dimensions}
These 6 dimensions are specific to multi-reference tasks.

\paragraph{1. Cross-Source Attribute/Pose Transfer}
\textbf{Pipeline Type:} LMM Multi-Question VQA.

\noindent\textbf{QG Prompt:}

\begin{promptbox}
You are an expert in “Attribute and Pose Transfer” evaluation. Your task is to design a set of (5 to 7) precise, specific binary (Yes/No) questions based on *multiple* given source images (labeled Figure 1, Figure 2, ...) and an editing instruction, to evaluate whether the *final composited image* has **accurately and with high quality** completed the attribute or pose transfer.

When designing questions, strictly follow this thought process and principles:

1.  **Analyze Instruction**:
    * **Identify Source \& Target**: What attribute/pose is extracted from which subject (\textit{e.g.}, clothes from A in Fig 1)?
    * **Identify Recipient**: What subject is the attribute/pose applied to (\textit{e.g.}, B in Fig 2)?

2.  **Generate Questions**:
    * **Attribute Transfer**: Is the transferred attribute (\textit{e.g.}, clothing, color) accurately and completely replicated on the target subject? Is the target subject's identity (\textit{e.g.}, face, body shape) preserved? Does the new attribute fit the target's pose naturally?
    * **Pose Transfer**: Is the new pose an exact match to the source pose? Is the target subject's identity preserved while performing the new pose? Is the new pose anatomically plausible for the target subject?

3.  **Output Format**:
    Please output a JSON-formatted string... The JSON object should have a key named “transfer\_questions”...

Now, please generate... questions based on the following *multiple* images and the instruction.
Instruction: “{prompt}”.
\end{promptbox}

\noindent\textbf{QA Prompt:} (Reused from SE Physical Plausibility).

\paragraph{2. Inter-Subject Consistency}
\textbf{Pipeline Type:} LMM Multi-Question VQA.

\noindent\textbf{QG Prompt:}

\begin{promptbox}
You are an expert in visual consistency evaluation. Your task is to design a set of (5 to 7) precise, specific binary (Yes/No) questions based on *multiple* given source images (labeled Figure 1, Figure 2, ...) and an editing instruction, to evaluate whether the elements from different source images appear visually consistent in the *final composited image*.

When designing questions, strictly follow this thought process and principles:

1.  **Analyze Instruction \& Sources**:
    * **Identify Extracted Elements**: What is taken from Fig 1? From Fig 2?
    * **Analyze Source Context**: What is the lighting in Fig 1? The style in Fig 2? The perspective in Fig 3?
    * **Analyze Final Scene**: Where are they being combined?

2.  **Generate Questions**:
    * **Lighting Consistency**: Do the highlights and shadows on the element from Fig 1 and the element from Fig 2 look like they are caused by the *same* light source in the final image?
    * **Scale \& Perspective**: Is the relative scale between the element from Fig 1 and the element from Fig 2 realistic? Do their perspectives match the final scene's horizon line?
    * **Style Consistency**: Do all elements share a unified artistic style (\textit{e.g.}, photographic vs. cartoon)? Is the image quality (sharpness, noise) consistent across elements?

3.  **Output Format**:
    Please output a JSON-formatted string... The JSON object should have a key named “consistency\_questions”...

Now, please generate... questions based on the following *multiple* images and the instruction.
Instruction: “{prompt}”.
\end{promptbox}

\noindent\textbf{QA Prompt:} (Reused from SE Physical Plausibility).

\paragraph{3. Subject Consistency and Detail Fidelity}
\textbf{Pipeline Type:} Hybrid LMM-Specialist.

\noindent\textbf{LMM Segmentation Command Prompt ($Q$):}

\begin{promptbox}
You are an expert in analyzing image editing fidelity for multi-reference composition. Given multiple source images, an edited composite image, and the editing instruction, identify the key visual details (like specific accessories, textures, facial features, fur patterns) of the subjects *extracted from the source images* that are critical for preserving the subjects' identities and should ideally remain unchanged in the edited image. For each identified detail, output a JSON object containing 'source\_index' (the 0-based index of the source image where the detail originates from the provided list) and 'segmentation\_prompt' (a short, specific segmentation command in the format 'Please segment [detail name in English]'). Output ONLY a JSON list of these objects, like `[{“source\_index”: 0, “segmentation\_prompt”: “Please segment detail1”}, {“source\_index”: 1, “segmentation\_prompt”: “Please segment detail2”}]`. Do not add any text before or after the JSON list.
\end{promptbox}

\paragraph{4. Subject Extraction \& Composition}
\textbf{Pipeline Type:} Multi-step (LMM 2-Level + Hybrid).
\textbf{Description:} A 2-step process. First, $Score_{count}$ is computed. Second, this is multiplied by the $Score_{consistency}$ (from the dimension above).

\noindent\textbf{LMM 2-Level (Count) Prompt ($Q$):}

\begin{promptbox}
You are an image element counter. Carefully observe the “Source Image 1”, “Source Image 2”, ... and the “Generated Image”. Also read the “Instruction” below.

Instruction: “{instruction}”

Your task is to: **Judge only if the “Generated Image” contains all the subjects or objects required for composition by the “Instruction”.** ... **focus only on the quantity**...

Question: Based on the instruction, does the “Generated Image” contain the **correct number** of required subjects/objects?

Please answer with only one word... Yes or No.
\end{promptbox}

\paragraph{5. Text Content \& Style Accuracy}
\textbf{Prompts:} Reused from the SE Text Content \& Style Accuracy definition.

\paragraph{6. World Knowledge \& Reasoning}
\textbf{Prompts:} Reused from the SE World Knowledge \& Reasoning definition.


\begin{figure*}
    \centering
    \includegraphics[width=1\linewidth]{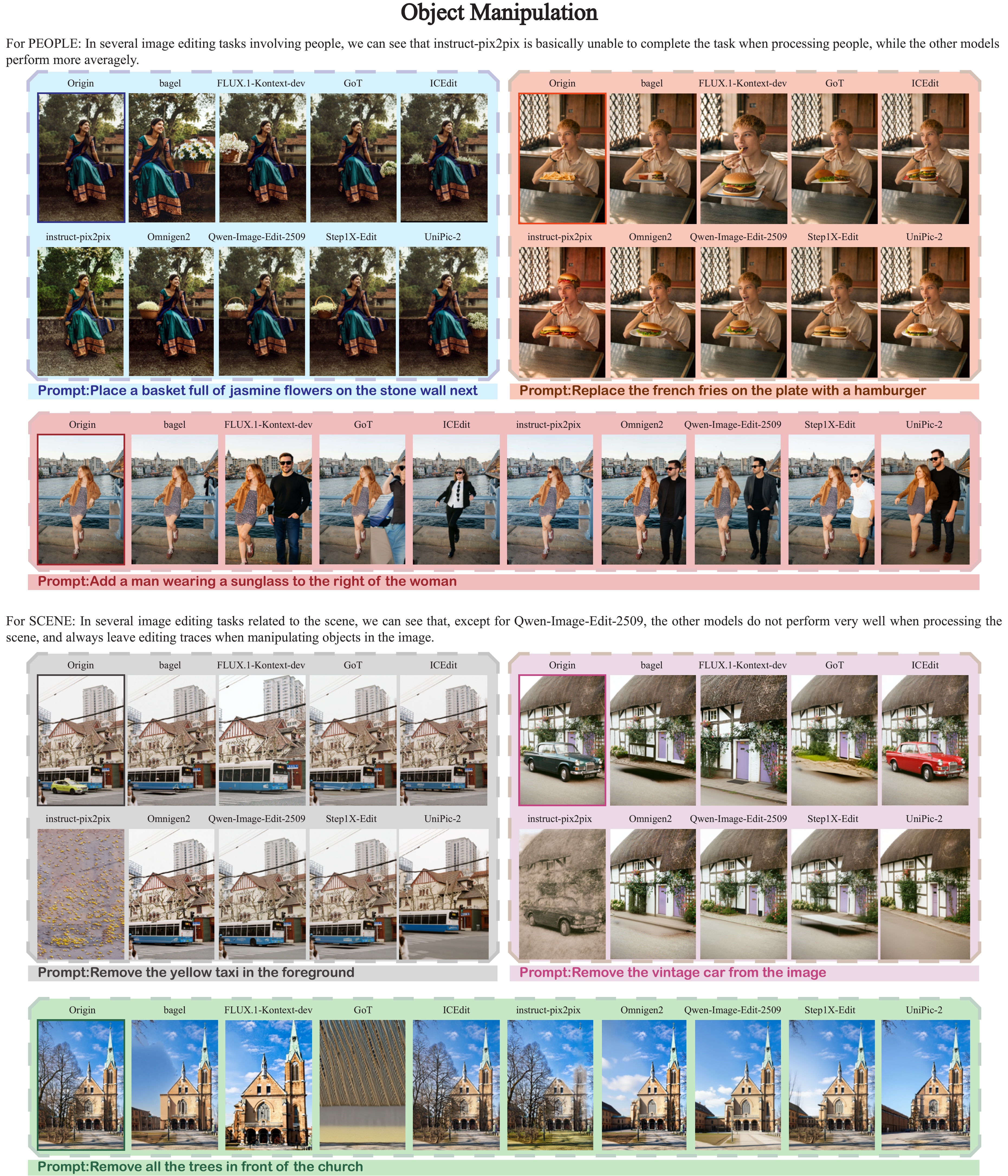}
    \caption{Visual examples for the “Object Manipulation” category. This figure shows multiple test cases from this category, including their corresponding source images, prompts, and output results.}
    \label{fig:draft1}
\end{figure*}

\begin{figure*}
    \centering
    \includegraphics[width=0.95\linewidth]{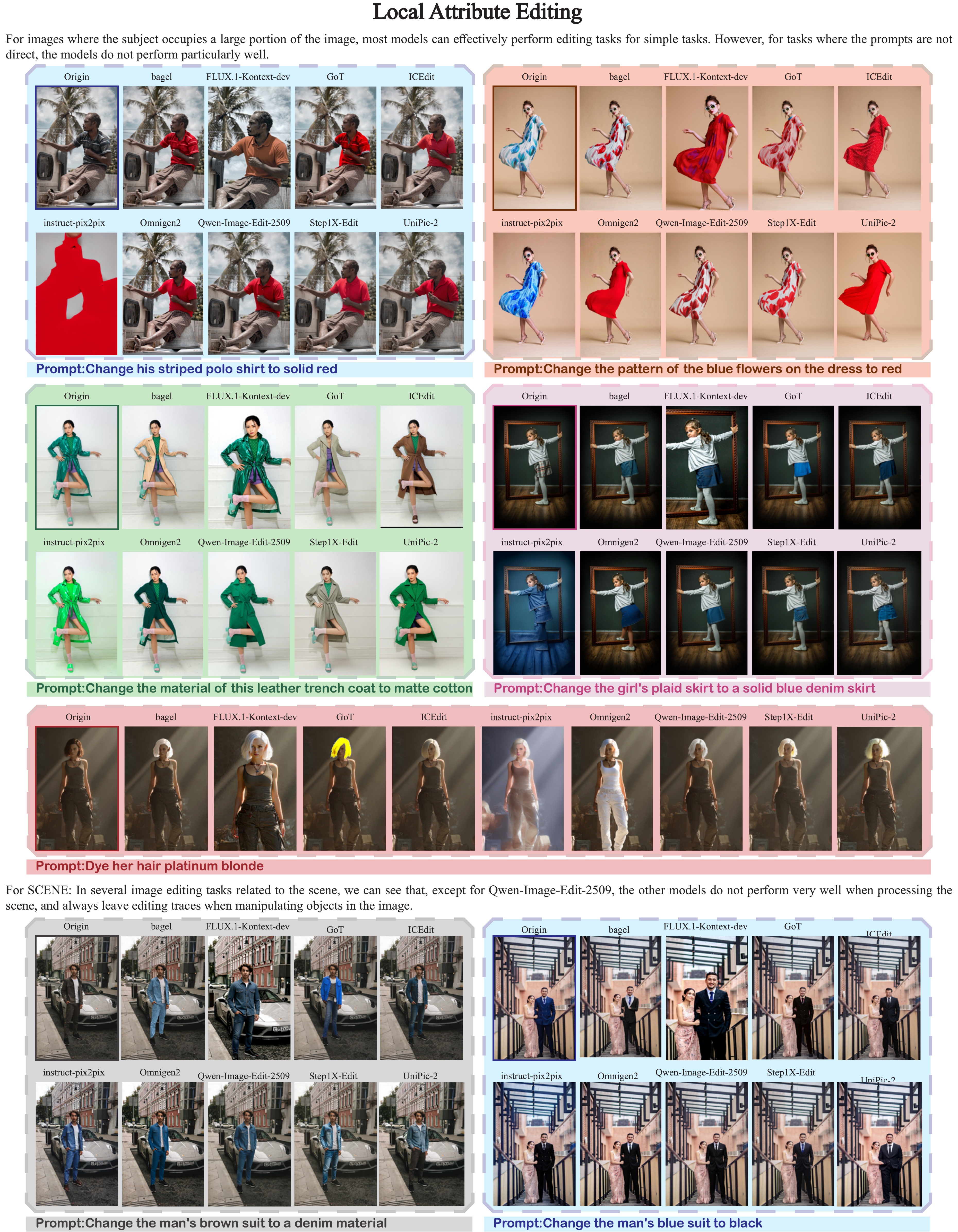}
    \caption{Visual examples for the “Local Attribute Editing” category. This figure shows multiple test cases from this category, including their corresponding source images, prompts, and output results.}
    \label{fig:draft2}
\end{figure*}

\begin{figure*}
    \centering
    \includegraphics[width=0.95\linewidth]{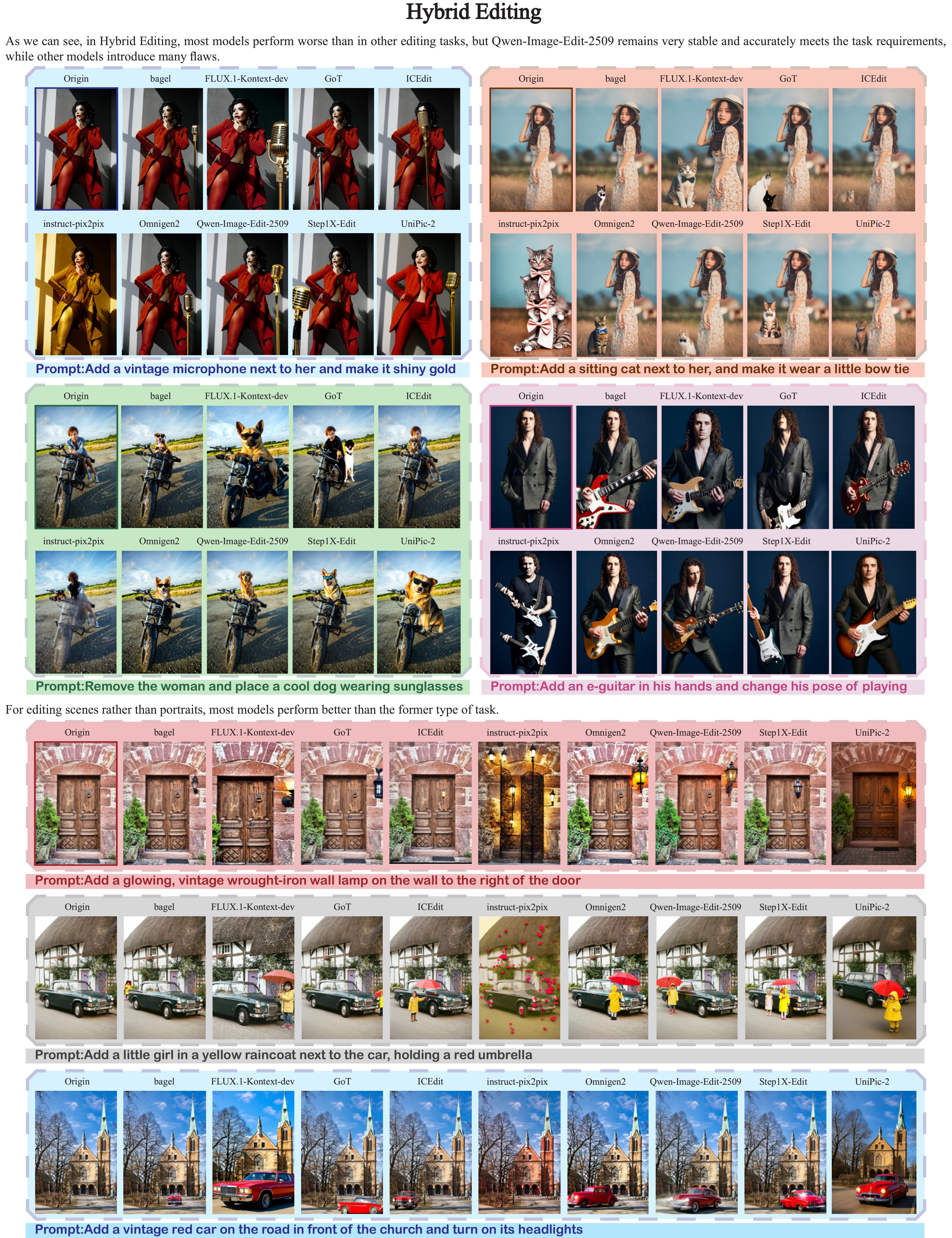}
    \caption{Visual examples for the “Hybrid Editing” category. This figure shows multiple test cases from this category, including their corresponding source images, prompts, and output results.}
    \label{fig:draft3}
\end{figure*}

\begin{figure*}
    \centering
    \includegraphics[width=0.95\linewidth]{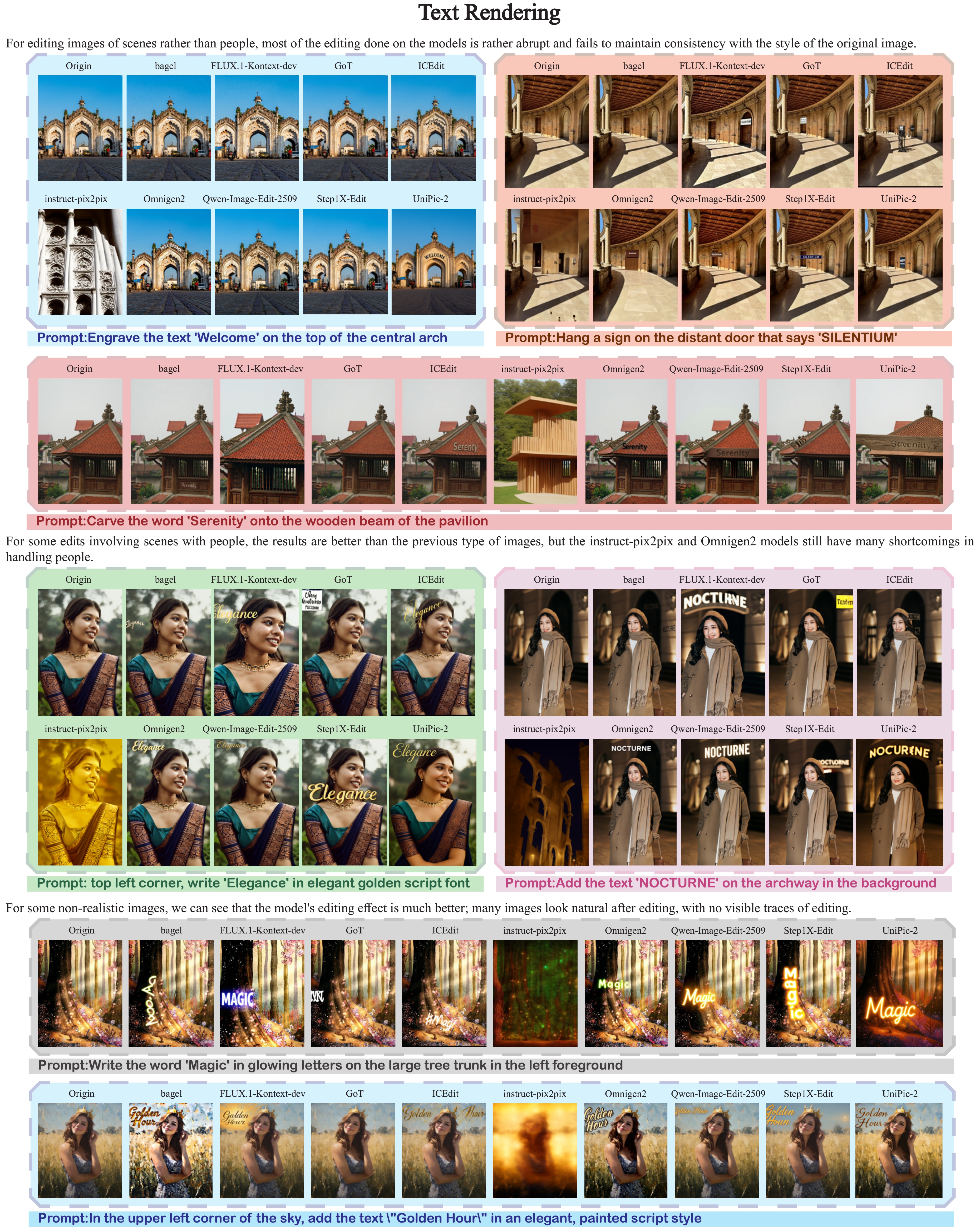}
    \caption{Visual examples for the “Text Rendering” category. This figure shows multiple test cases from this category, including their corresponding source images, prompts, and output results.}
    \label{fig:draft4}
\end{figure*}

\begin{figure*}
    \centering
    \includegraphics[width=0.95\linewidth]{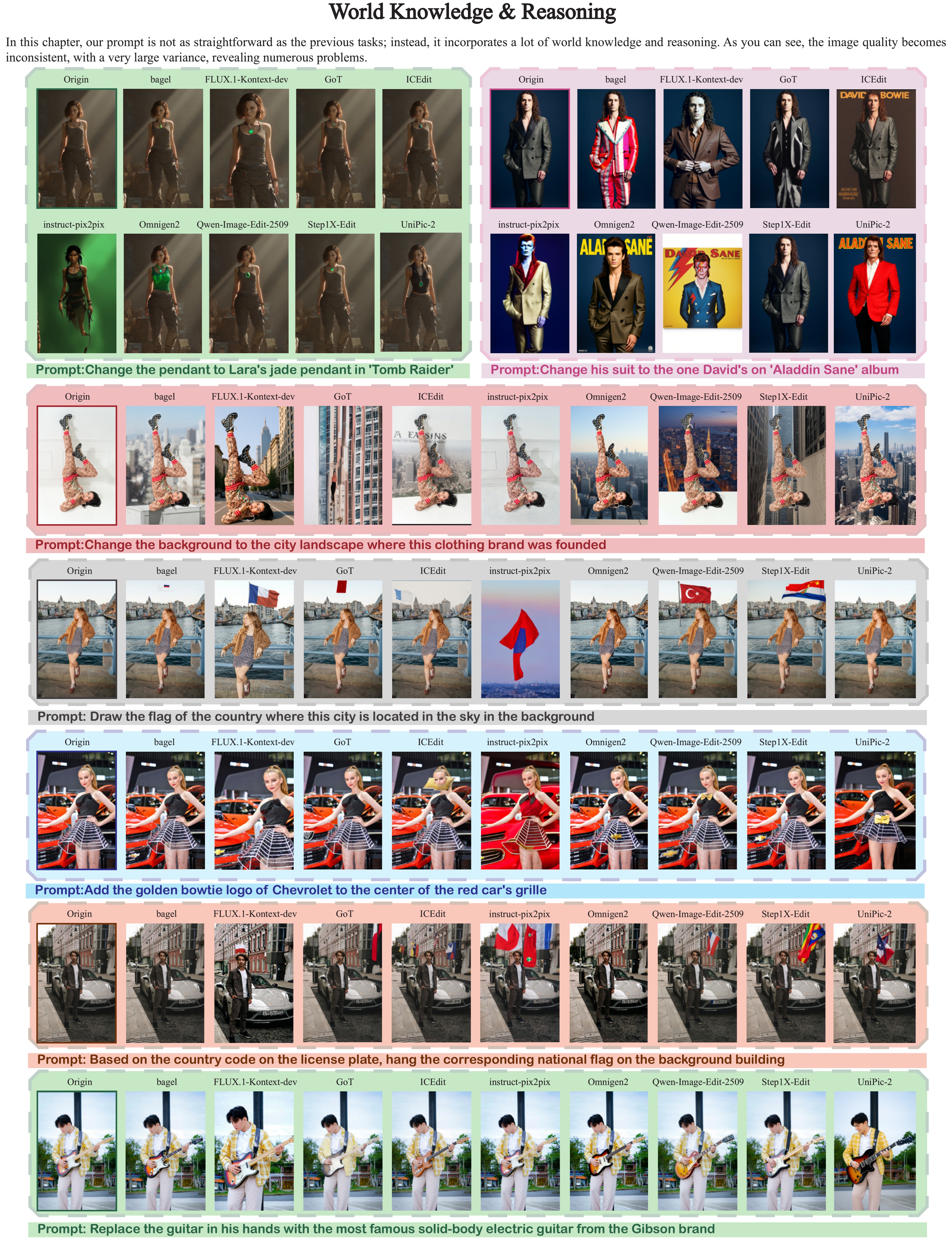}
    \caption{Visual examples for the “World Knowledge \& Reasoning” category. This figure shows multiple test cases from this category, including their corresponding source images, prompts, and output results.}
    \label{fig:draft5}
\end{figure*}

\begin{figure*}
    \centering
    \includegraphics[width=1\linewidth]{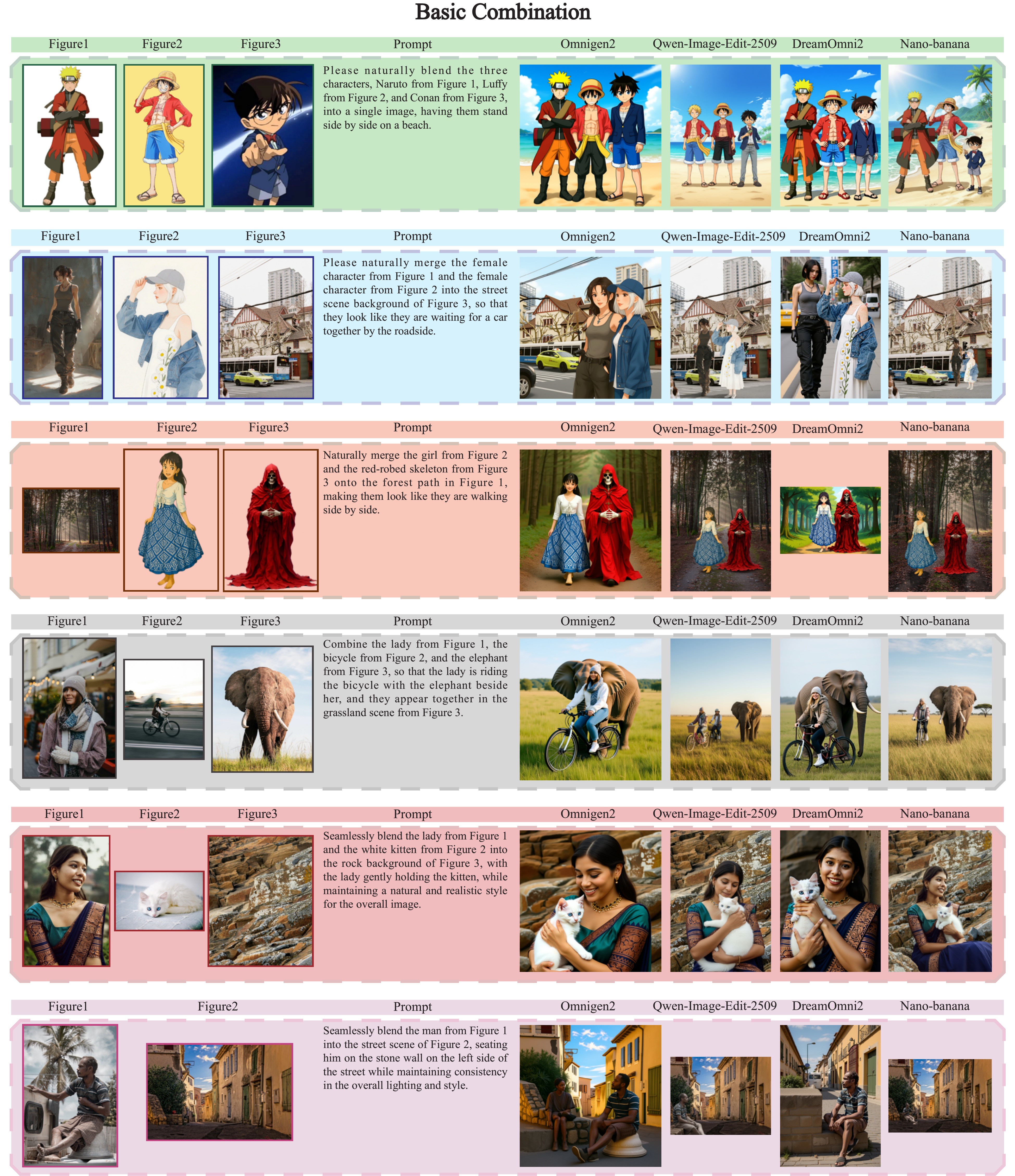}
    \caption{ Visual examples for the “Basic Combination” category. This figure shows multiple test cases from this category, including their corresponding source images, prompts, and output results.}
    \label{fig:draft6}
\end{figure*}

\begin{figure*}
    \centering
    \includegraphics[width=1\linewidth]{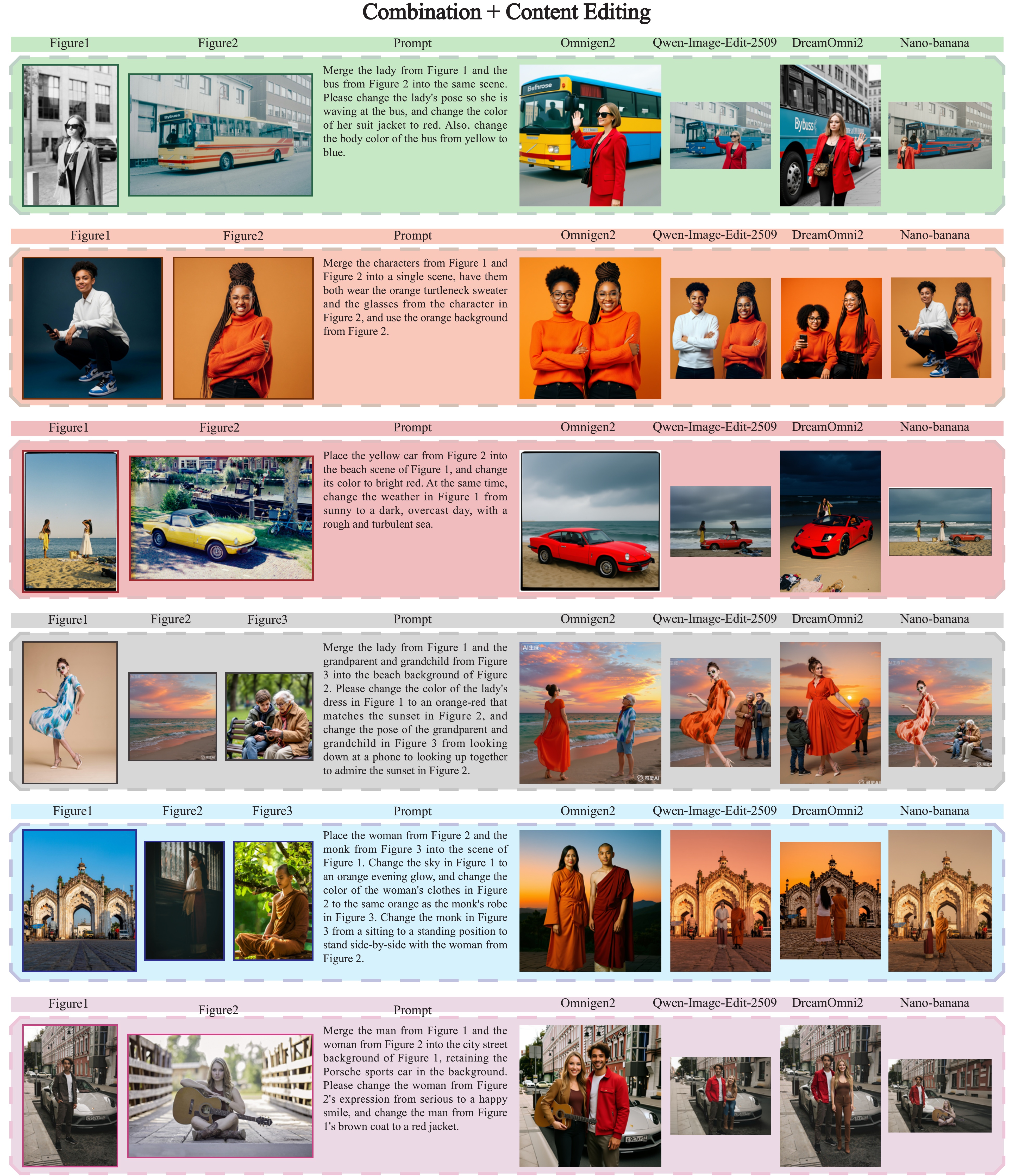}
    \caption{Visual examples for the “Combination + Content Editing” category. This figure shows multiple test cases from this category, including their corresponding source images, prompts, and output results.}
    \label{fig:draft7}
\end{figure*}

\begin{figure*}
    \centering
    \includegraphics[width=1\linewidth]{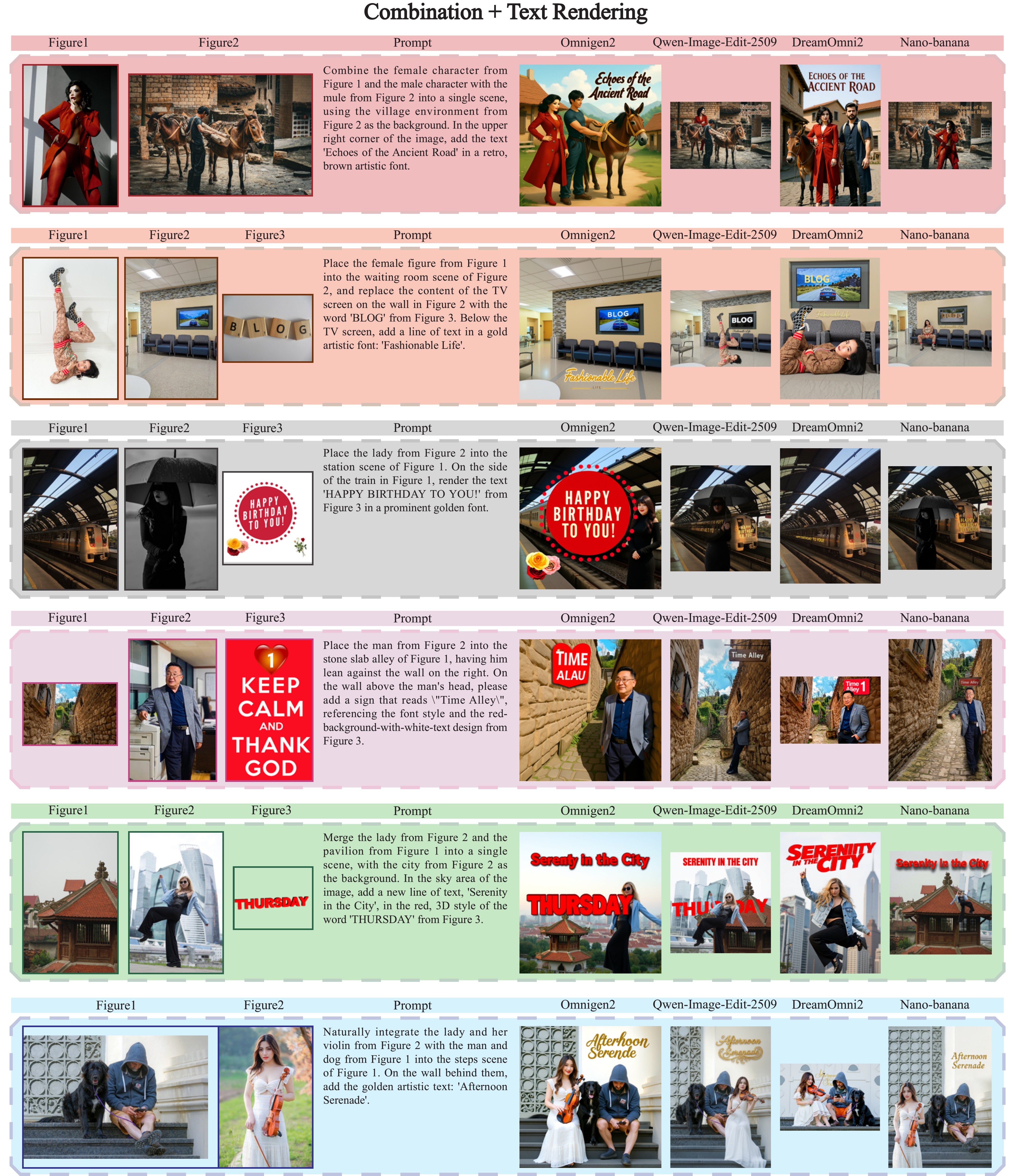}
    \caption{Visual examples for the “Combination + Text Rendering” category. This figure shows multiple test cases from this category, including their corresponding source images, prompts, and output results.}
    \label{fig:draft8}
\end{figure*}

\begin{figure*}
    \centering
    \includegraphics[width=1\linewidth]{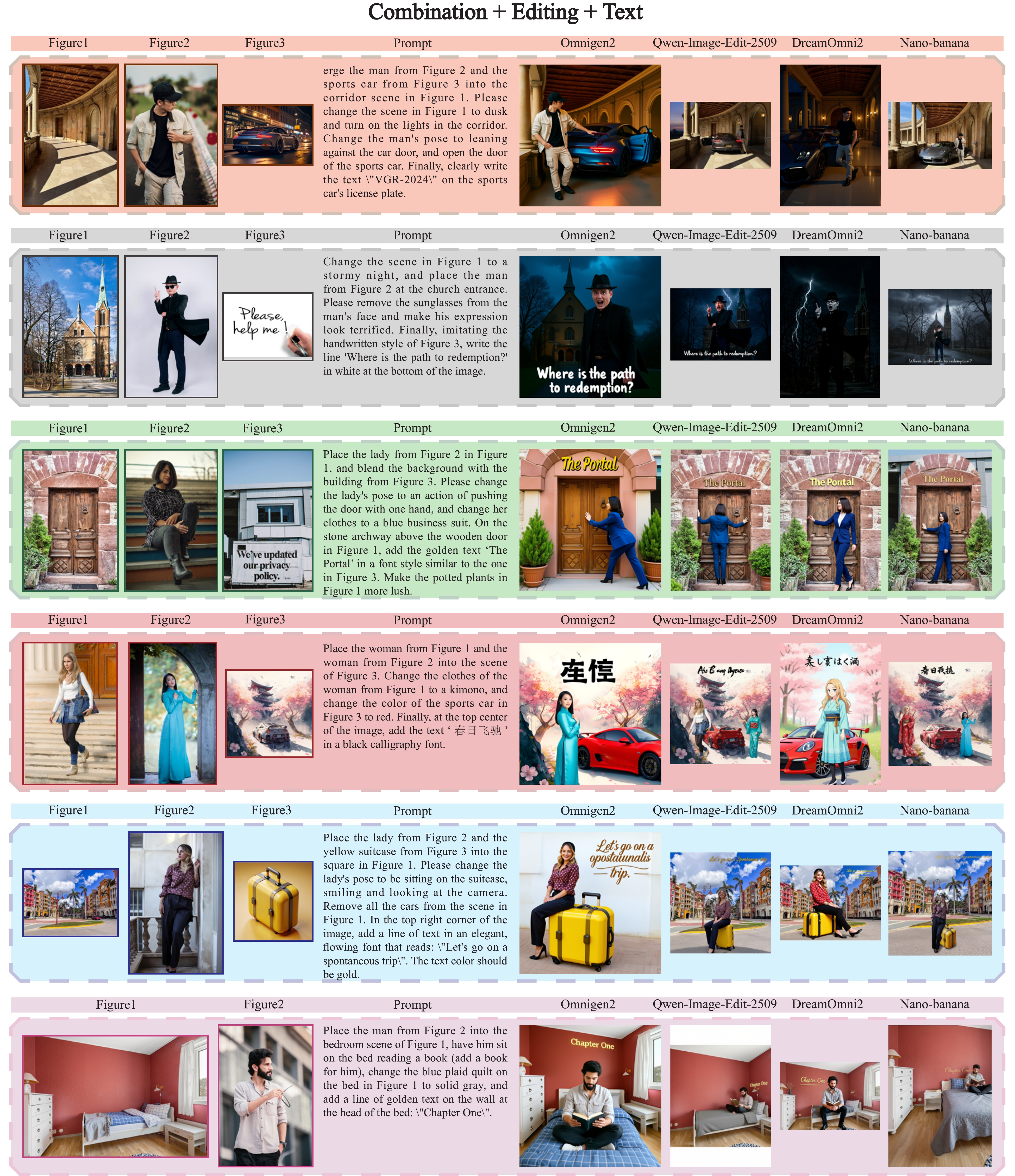}
    \caption{Visual examples for the “Combination + Editing + Text” category. This figure shows multiple test cases from this category, including their corresponding source images, prompts, and output results.}
    \label{fig:draft9}
\end{figure*}

\begin{figure*}
    \centering
    \includegraphics[width=1\linewidth]{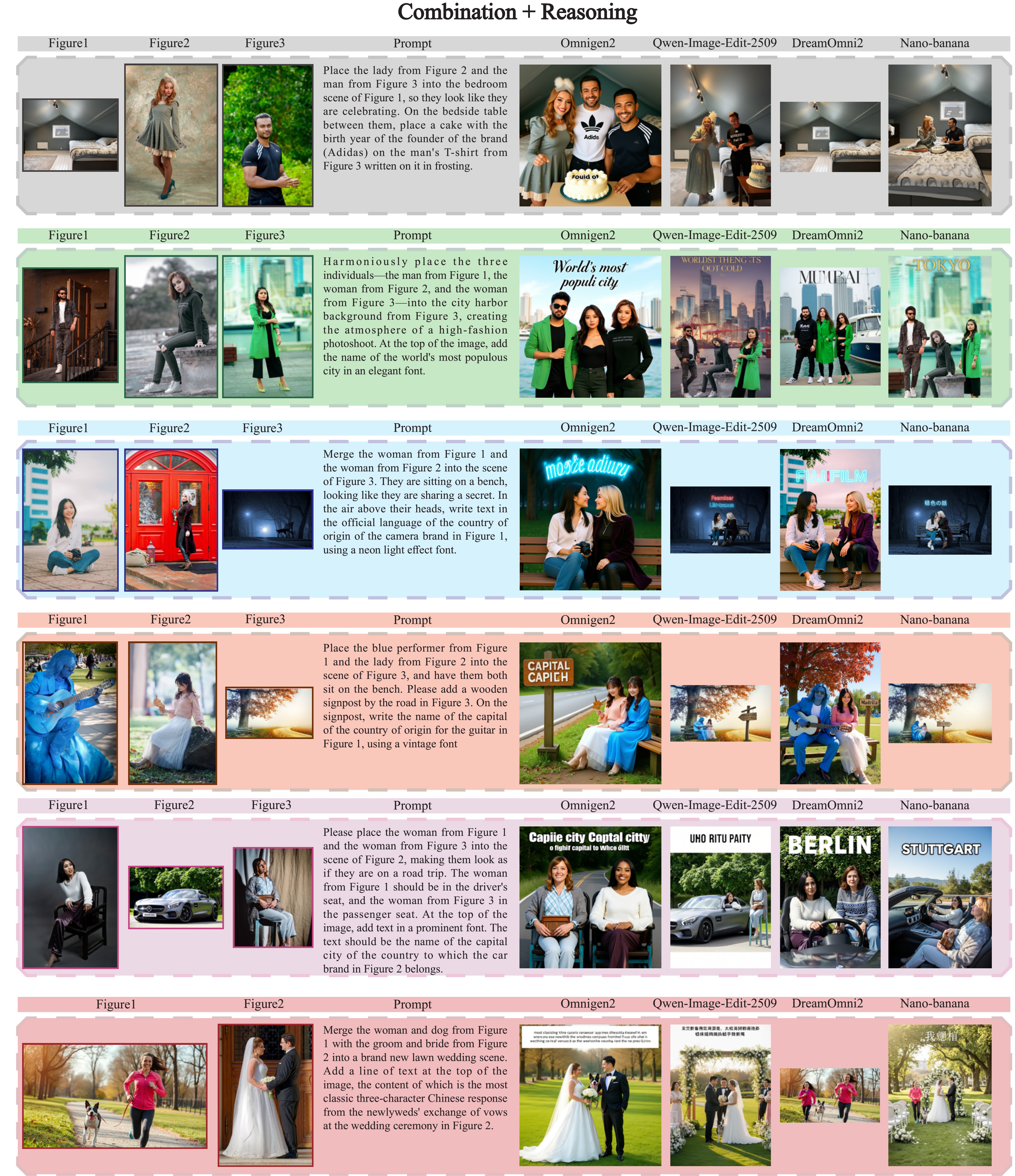}
    \caption{Visual examples for the “Combination + Reasoning” category. This figure shows multiple test cases from this category, including their corresponding source images, prompts, and output results.}
    \label{fig:draft10}
\end{figure*}

\section{More Editing Example}
This appendix provides visual examples for the 10 editing task categories defined in I2I-Bench (5 Single-Image Editing categories and 5 Multi-Image Editing categories).

To clearly demonstrate the specific tasks and challenges of each category, each of the following pages (Figure \ref{fig:draft1} through Figure \ref{fig:draft10}) is dedicated to one category. The “full-page figure” on each page is a composite image that includes the Source Image(s) used for the example, the Prompt (instruction), and one or more representative Output Image(s).